\documentclass[10pt,twocolumn,letterpaper]{article}
\pdfoutput=1
\usepackage{cvpr}
\usepackage{times}
\usepackage{epsfig}
\usepackage{graphicx}
\usepackage{amsmath}
\usepackage{amssymb}

\usepackage[breaklinks=true,bookmarks=false]{hyperref}

\usepackage{epstopdf}
\usepackage{lipsum}
\usepackage{tabulary}
\usepackage{tabularx}
\usepackage{multirow}
\usepackage{subcaption}
\usepackage{footmisc}
\usepackage{verbatim}
\usepackage{color}
\usepackage{makecell}
\usepackage{hhline}
\usepackage{arydshln}
\usepackage{kotex}

\usepackage{array}
\usepackage[skip=2pt]{caption}
\newcolumntype{C}[1]{>{\centering\let\newline\\\arraybackslash\hspace{0pt}}m{#1}}

\cvprfinalcopy 

\def\cvprPaperID{2900} 

\begin{document}

\title{Residual Features and Unified Prediction Network for Single Stage Detection}

  \author{Kyoungmin Lee \hspace{0.1in} Jaeseok Choi \hspace{0.1in} Jisoo Jeong \hspace{0.1in} Nojun Kwak\\
Seoul National University \\
{\tt\small \{strx2322, jaeseok.choi, soo3553, nojunk\}@snu.ac.kr}
}

\maketitle

\begin{abstract}

Recently, a lot of single stage detectors using multi-scale features have been actively proposed. They are much faster than two stage detectors that use region proposal networks (RPN) without much degradation in the detection performances.   
However, the feature maps in the lower layers close to the input which are responsible for detecting small objects in a single stage detector have a problem of insufficient representation power because they are too shallow. 
There is also a structural contradiction that the feature maps have to deliver low-level information to next layers as well as contain high-level abstraction for prediction.
In this paper, we propose a method to enrich the representation power of feature maps using Resblock and deconvolution layers. In addition, a unified prediction module is applied to generalize output results
and boost earlier layers' representation power for prediction.
The proposed method enables more precise prediction, which achieved higher score than SSD on PASCAL VOC and MS COCO. In addition, it maintains the advantage of fast computation of a single stage detector, which requires much less computation than other detectors with similar performance.
\end{abstract}


\section{Introduction}
The development of deep neural networks (DNN) in recent years has achieved remarkable results not only in object detection but also in many other areas. In the early researches of object detection using DNN, much attention has been paid to representation learning that can replace hand-crafted features without much consideration on the speed of detectors. Recently, real-time detectors with low computational complexities have been actively researched.

Researches on two-stage detectors, mostly based on Faster R-CNN \cite{ren2015faster}, applied the \textit{region proposal network} (RPN) and RoI pooling to the feature maps extracted by a state-of-the-art classifier, such as ResNet-101 \cite{he2016deep}.
On the other hand, the single-stage methods such as YOLO \cite{redmon2016you} and SSD \cite{liu2016ssd} removed RoI pooling layer and predict bounding boxes and corresponding class confidences directly while enabling faster detection and end-to-end learning.

\begin{figure}[t]
    \begin{subfigure}[b]{\linewidth}
    \includegraphics[width=0.49\linewidth]{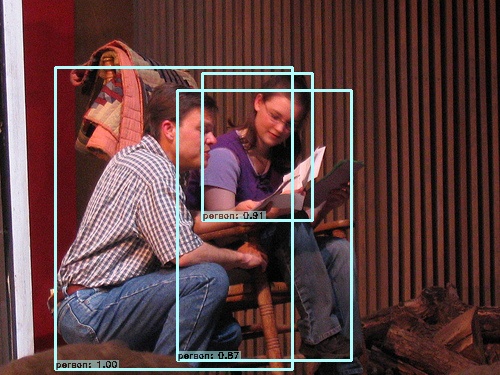}\hfill
    \includegraphics[width=0.49\linewidth]{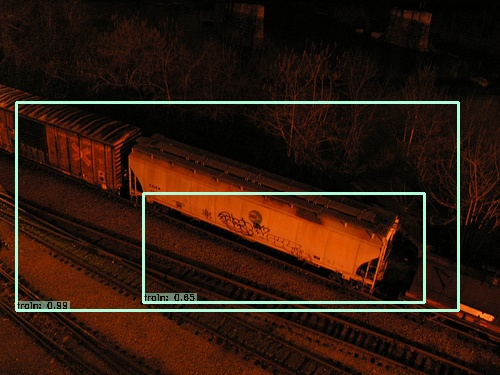}  \\
    \vspace{-0.3cm}
    \end{subfigure}
    \begin{subfigure}[b]{\linewidth}
    
    \includegraphics[width=0.49\linewidth]{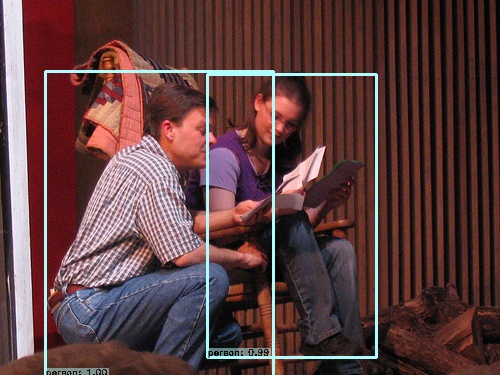}\hfill
    \includegraphics[width=0.49\linewidth]{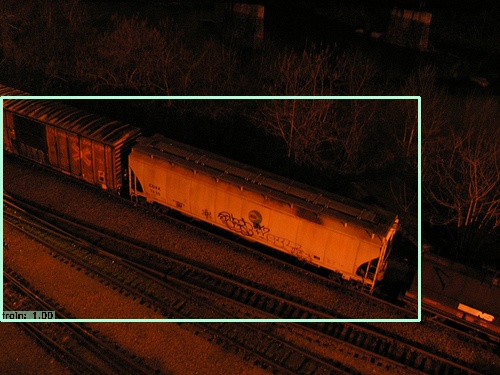}  \\
    
  \end{subfigure}
   \caption{
    \textbf{Box-in-Box problem}. Top: SSD300. Bottom: RUN300 (proposed). SSD detects objects with overlapping boxes which are redundant.}
    \label{fig:boxinbox}
\end{figure}

Especially SSD makes use of multi-scale feature maps generated from a backbone network such as VGG-16 \cite{simonyan2014very} to detect objects in various sizes. Since each of the prediction modules composed of 3 $\times$ 3 convolution filters detects bounding box on each layer separately, they cannot reflect appropriate contextual information from different scales. It causes the problem named as ``Box-in-Box'' \cite{jeong2017enhancement} as shown in Figure \ref{fig:boxinbox}. 
In the figure, we can see that SSD often detects a single object with two overlapping boxes. The smaller box has partial image such as the upper body of a person or the head of an animal. 

To solve the problem, \cite{fu2017dssd,lin2017focal} used ResNet and \textit{feature pyramid network} (FPN) \cite{lin2016feature} structure to inject larger contextual information through deep convolutional back-bone by the use of deconvolution. However, these structures have the disadvantage of increasing the computational complexity, thus reduces detection speed, which is a key advantage of a single-stage detector.

In this paper, we propose very simple ideas to solve the essential problems of multi-scale single stage detectors. First, we introduce a 3-way residual block, which is a structure where the Resblock \cite{he2016deep} and the deconvolution layer are added on the multi-scale feature maps. It makes detected boxes be determined with larger context and be more reliable. Second, we integrate the multiple prediction modules, which had been applied separately to each layer, into one to boost information level of feature maps from earlier layers.

The proposed structure, called \textit{``RUN; Residual features and Unified prediction Network"}, is a single-stage detector that combines 3-way Resblock with unified prediction module on VGG-16 network. RUN is not only very compact and fast compared to other ResNet-based two-stage detectors and single-stage detectors using FPN, but it also achieves superior or competitive performance compared to other competitors.

\begin{figure*}[ht]
\begin{center}
	\begin{subfigure}{\textwidth}
        \centering
    	\includegraphics[height=5cm]{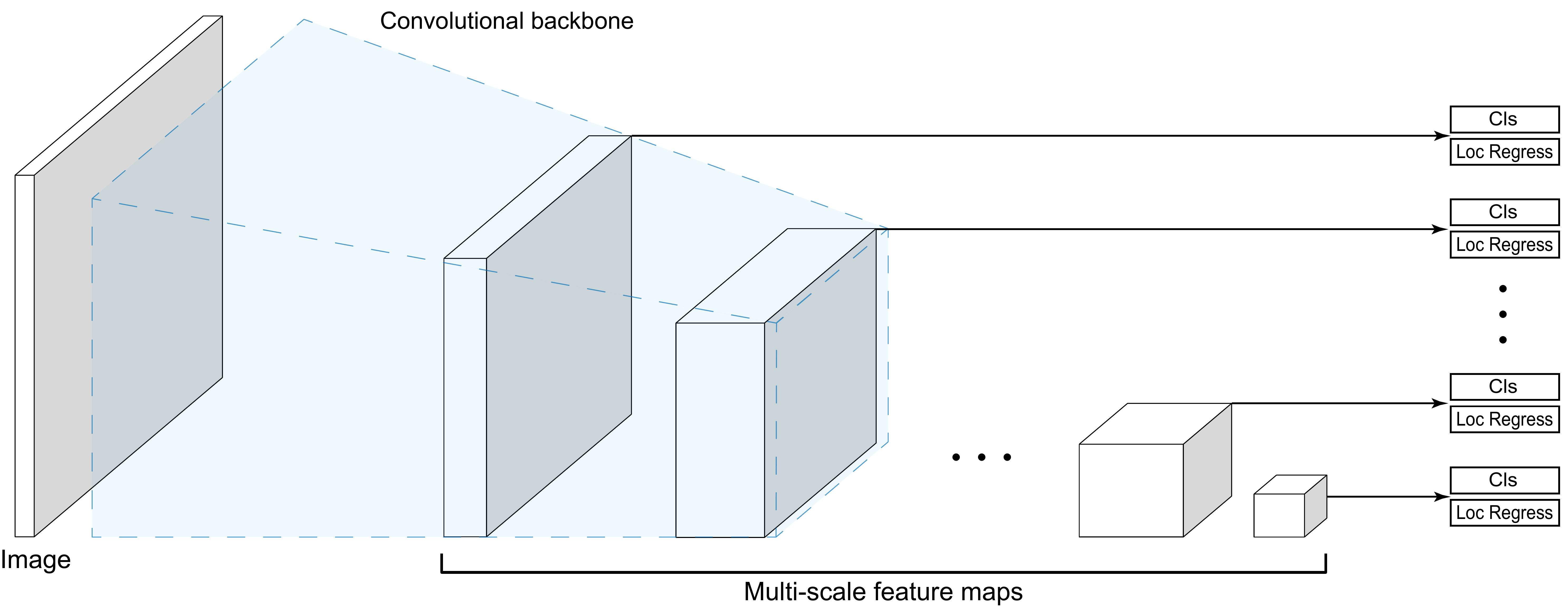}
        \vspace{5mm} 
	\end{subfigure}
	\begin{subfigure}{\textwidth}
    	\centering
    	\includegraphics[height=5cm]{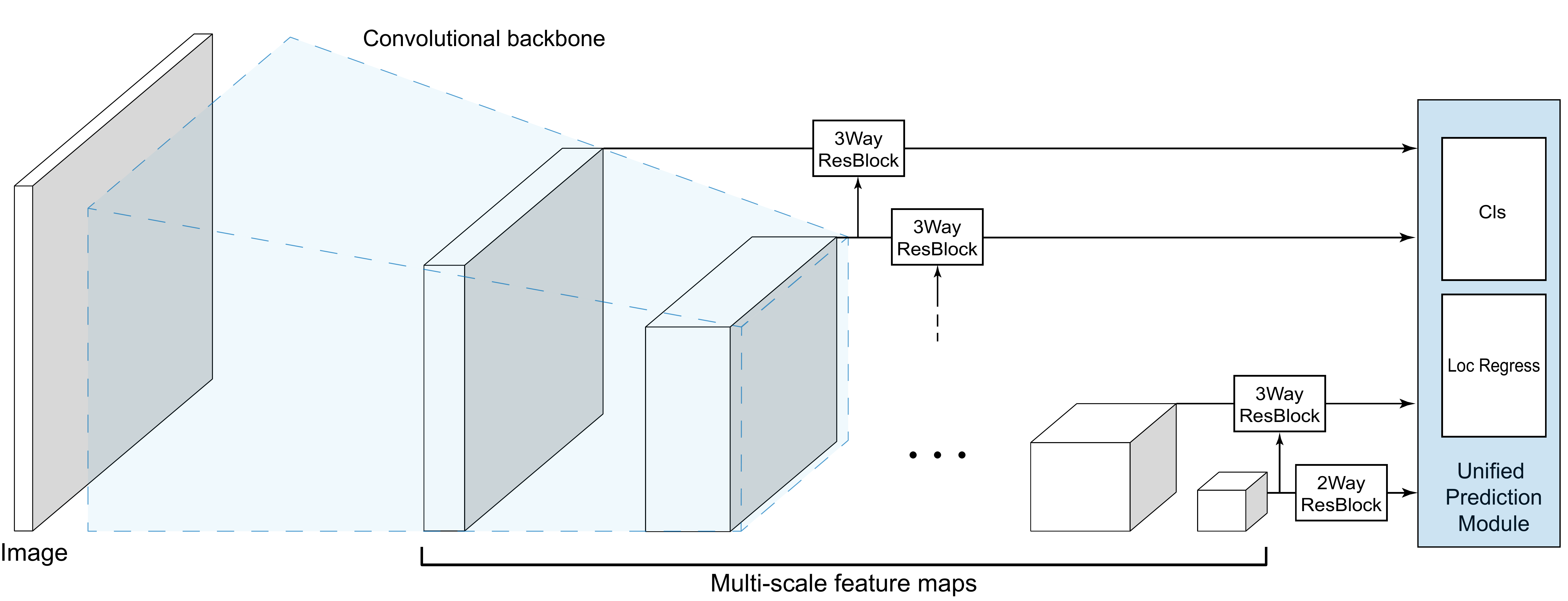}
	\end{subfigure}
\end{center}
   \caption{
    \textbf{Networks of SSD and RUN}. Top: SSD. Bottom: RUN. Compared with SSD, RUN has residual blocks and unified prediction module. The arrow from the bottom to the top indicates the deconvolution branch.}
   \label{fig:arch} 
\end{figure*}

\section{Related Works}
\label{sec:related_work}
Overfeat \cite{sermanet2013overfeat}, SPPNet \cite{he2014spatial}, R-CNN \cite{girshick2014rich}, Fast R-CNN \cite{girshick2015fast}, Faster R-CNN  \cite{ren2015faster} and R-FCN \cite{li2016r} which are classified as \textit{region-based convolutional neural networks} (R-CNN) showed a tremendous improvement in performance compared to the previous object detection techniques. These region-based approaches have achieved huge advances over the last few years and are still the state-of-the-art approaches among many object detection techniques. Specifically these approaches 
usually use a two-stage method of generating a number of bounding boxes and then assigning a classification score to the bounding boxes. Thus, although classification may be relatively accurate, these are too slow to be used for real-time applications.

Redmon \etal \cite{redmon2016you} proposed a method named as YOLO to predict bounding boxes and associated class probabilities in a single step by framing object detection as a regression problem. It divides input images to grid maps and regresses bounding boxes for multiple objects on each grid. This was the beginning of single stage detection and subsequently inspired structures such as SSD \cite{liu2016ssd}. However, since YOLO uses only the highest-level{\footnote{In this paper, the term \textit{level} is used interchangeably with \textit{layer}. Highest level indicates the the farthest layer from the input layer.} feature maps to detect objects, there is a lack of lower-level information, which results in somewhat inaccurate detection, especially for small objects.

In order to solve this problem, SSD \cite{liu2016ssd} utilized not only the highest-level features but also lower-level features which have enough resolution to detect small objects. As mentioned in Inside-Outside Net (ION) \cite{bell2016inside} and HyperNet \cite{kong2016hypernet}, each feature maps at different layers have different abstraction levels for an input image. Therefore, it is clear that using multi-scale feature maps can improve detection performance for objects of various scales. In SSD, many default boxes are created in the feature maps and bounding box regression and classification are performed for each box area using 3$\times$ 3 convolution. This method enables multi-scale object detection without using RoI pooling. In addition, it can effectively improve the detection accuracy of small objects which is a disadvantage of YOLO \cite{redmon2016you}.
However, as mentioned in MS-CNN \cite{cai16mscnn}, SSD has the problem that back-propagation allows the gradient to cause unnecessary deformations in the feature maps since the feature maps of the backbone network are used directly in bounding box regression and classification. Then, it can lead to some instability during learning. In addition, since each classifier only uses single scale feature maps, it cannot reflect larger or smaller contextual information other than the one for the corresponding scale. 

Recently, various methods have attempted to enhance the contextual information of each layer while taking advantage of SSD \cite{liu2016ssd}. DSSD \cite{fu2017dssd} could obtain higher accuracy by changing the base network to ResNet-101 \cite{he2016deep} and combining the FPN \cite{lin2016feature} using deconvolution layers in combination with the existing multiple layers to reflect the large scale context. However, with the use of deep structure of ResNet-101 and deconvolution layers, the processing speed degrades much (under 16.4 images per second), which prohibits the method to be used for real-time detection problems. 

Ren \etal \cite{Ren17CVPR} introduced a \textit{recurrent rolling convolution (RRC)} architecture to improve detection performance by mutually complementing layers having different sizes of contextual information. RRC made multi-scale feature maps include large and small context by concatenating adjacent feature maps by pooling and deconvolution. This process was implemented by RNN structure and it allowed to reflect not only the information of the adjacent feature maps but also the information of the remote feature maps. 

Unlike RRC \cite{Ren17CVPR}, Rainbow SSD (R-SSD) \cite{jeong2017enhancement} proposed a method to concatenate feature maps not only in the adjacent layers but also in all the layers for bounding box regression and classification using pooling and deconvolution. 
It achieves higher performance than SSD by enhancing representation power of feature maps. Also, by making the dimension of each layer the same, it made it possible to use a unified prediction module instead of different prediction modules for different layers. Woo \etal \cite{woo2017stairnet} proposed StairNet which utilizes both FPN \cite{lin2016feature} structure of base VGG-16 network and unified prediction of R-SSD.

Additionally, Lin \etal \cite{lin2017focal} redefined the loss term for object detection which is named as Focal Loss. Unlike other batch reconstruction methods like OHEM \cite{shrivastava2016training}, it effectively resolves the foreground-background imbalance problem by changing the loss term. Their RetinaNet which uses Focal Loss in combination with ResNet and FPN \cite{lin2016feature} structure achieved the state-of-the-art performance.

\section{Residual Feature and Unified Prediction Network (RUN)}
\label{sec:RUN}

In this section, we propose \textit{residual feature maps} and \textit{unified prediction module}. It shows how the addition of a structurally simple idea can complement the drawbacks of SSD-based single-stage object detection methods.

\begin{figure*}[ht]
\begin{center}
	\begin{subfigure}{\textwidth}
    	\hspace{1.5cm}
		\includegraphics[height=4.5cm]{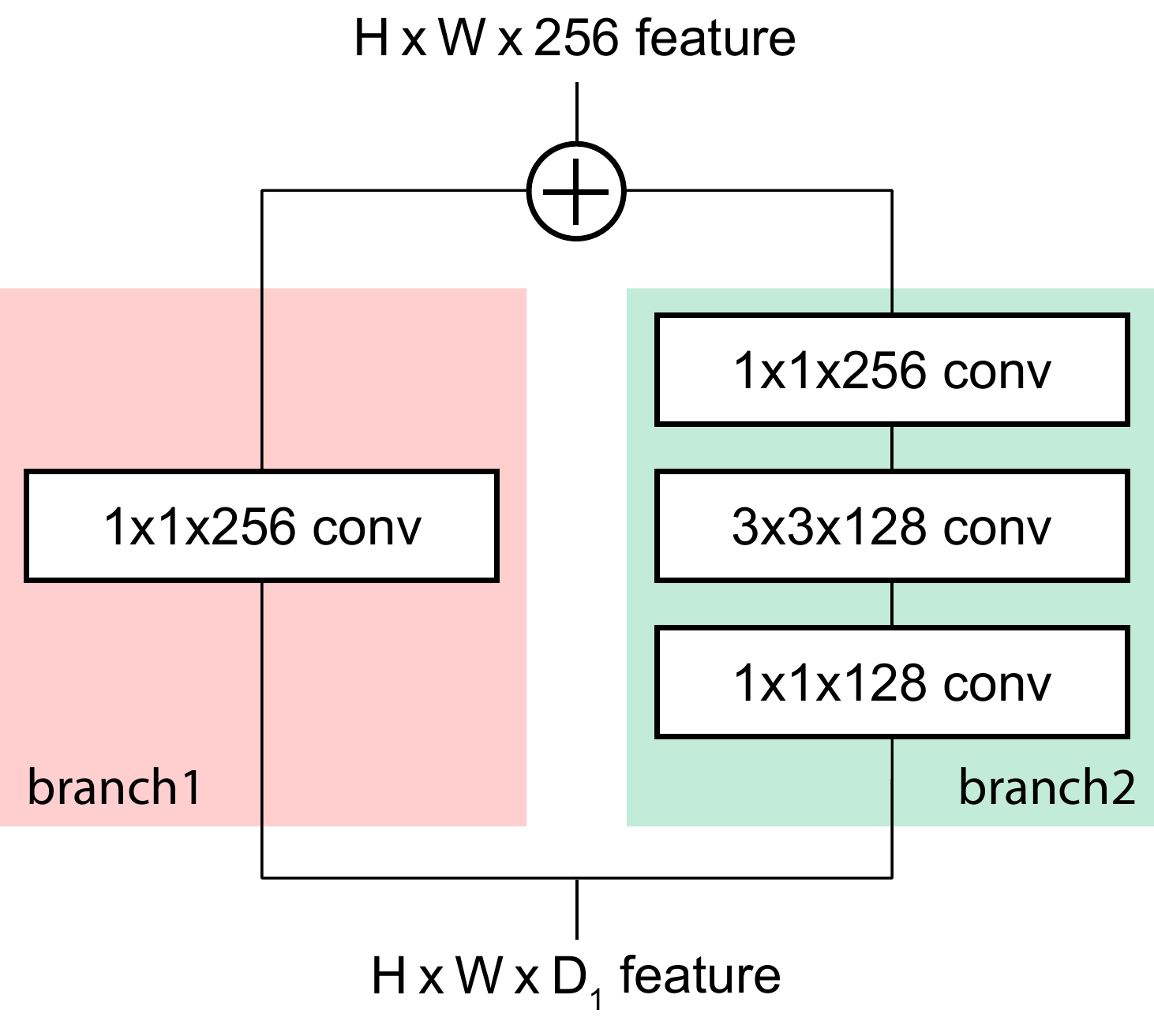}
        \hspace{2cm}
        \includegraphics[height=4.5cm]{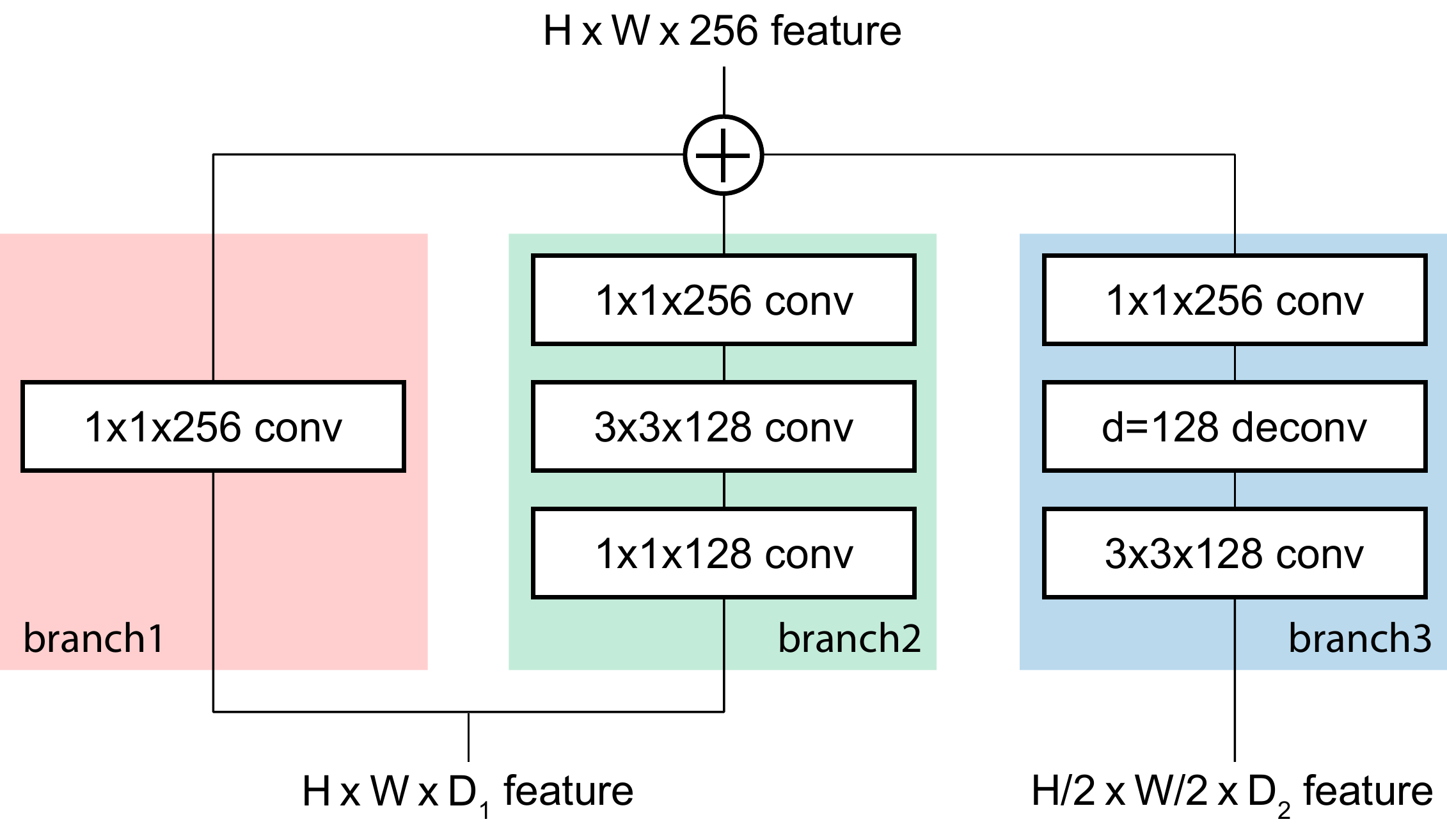}
	\end{subfigure}
\end{center}
\caption{
\textbf{Residual blocks}. Left: 2-way Resblock. Right: 3-way Resblock with deconvolution branch (branch3).}
\label{fig:Resblock}
\end{figure*}

\subsection{Residual Feature Maps} 
Recent CNN models designed for object detection makes use of a backbone network which is originally devised to solve image classification problems. 
Although the detection network can be trained end-to-end, the backbone network is normally initialized with the weights for the image classification problems.
The relation between the features and predictions in the networks used for image classification can be expressed mathematically as follows:
\begin{equation} \label{eq:3.1}
 \mathbf{x}_{n} = \mathcal{F}_{n}(\mathbf{x}_{n-1}) = (\mathcal{F}_{n} \circ \mathcal{F}_{n-1}  \circ \dotsm \circ \mathcal{F}_{1}) (\mathbf{I})
\end{equation}
\begin{equation} 
 \mathbf{Scores} = \mathcal{P}(\mathbf{x}_{n}),
\end{equation}
where $\mathbf{I}$ is an input image, $\mathbf{x}_{n}$ is the $n^{th}$-level feature map, $\mathcal{P}$ is a prediction function, and $\mathcal{F}_{n}$ is a combination of non-linear transformations such as convolution, pooling, ReLU, etc. Here, the top feature map, $\mathbf{x}_{n}$, learns information on high-level abstraction. On the other hand, $\mathbf{x}_{k}$ ($k < n$) has more local and low-level information as $k$ becomes smaller.

SSD \cite{liu2016ssd} applies several feature maps with different scales directly as an input to separate prediction modules to calculate object positions and classification scores, which can be denoted by the following equation:
\begin{equation} \label{eq:3.3}
 \mathbf{Detection} = \big\{ 
 \mathcal{P}_{1}(\mathbf{x}_{s_{1}}), 
 \mathcal{P}_{2}(\mathbf{x}_{s_{2}}),
 \dotsc, 
 \mathcal{P}_{k}(\mathbf{x}_{s_{k}})
 \big\},
\end{equation}
where $s_{1}$ to $s_{k}$ are feature indices for source feature maps for multi-scale prediction, $\mathcal{P}_k$ is a function that outputs multiple objects with different positions and scores. 
Combining \eqref{eq:3.1} and \eqref{eq:3.3}, it can be expressed as 
\begin{multline} \label{eq:3.4}
 \mathbf{Detection} = \big\{
 \mathcal{P}_{1}(\mathbf{x}_{s_{1}}), 
 \mathcal{P}_{2} \big( \mathcal{F}_{s_{1}}^{s_{2}} (\mathbf{x}_{s_{1}}) \big), \\
 \dotsc,
 \mathcal{P}_{k} \big( \mathcal{F}_{s_{1}}^{s_{k}} (\mathbf{x}_{s_{1}}) \big) 
 \big\},
\end{multline}
where $\mathcal{F}_{a}^{b} (\mathbf{x}_{a}) \triangleq 
(\mathcal{F}_{b} \circ \dotsm \circ \mathcal{F}_{a+1} ) (\mathbf{x}_{a}).$
Here, the earlier feature map $ \mathbf{x}_{s_{1}} $ needs to learn high-level abstraction to improve the performance of $ \mathcal{P}_{1}(\mathbf{x}_{s_{1}})$. At the same time, it also needs to learn local features for efficient information transfer to the next feature maps.
This not only makes learning difficult, but also causes the overall performance to decrease.

To resolve this problem, SSD \cite{liu2016ssd} added L2 normalization layer between the conv4\_3 layer and the prediction module, which results in a reduced magnitude of the gradients from the prediction module. Cai \etal\cite{cai16mscnn} tried to solve this problem by adding a convolution layer only to the conv4\_3 layer. Since the above problem is not solely on the conv4\_3 layer, the aforementioned approaches do not essentially solve the problem.
To meet this contradictory requirement of maintaining low-level information while having the flexibility to learn high-level abstraction, it is desired to separate and decouple the backbone network and the prediction module in the training phase.

In order to solve the same problem, we propose a new architecture that decouples backbone network from the prediction module as shown in Figure \ref{fig:arch}. Instead of directly connecting the feature maps in the backbone network to the prediction module, we inserted a multi-way Resblock for each level of feature maps, which acts like a bumper. The detailed architecture of the proposed multi-way Resblocks are shown in Figure \ref{fig:Resblock}.  
Convolution layers and nonlinear activation units are used for all branches of the proposed Resblock. This prevents the gradients of the prediction module from flowing directly into the feature maps of the backbone network. Also, it clearly distinguishes the features to be used for prediction from the features to be delivered to the next layer. 
In other words, the proposed Resblock takes the role of learning high-level abstraction for object detection, while the backbone network containing low-level features is designed to be intact from the high-level detection information.
This design helps to improve the feature structure of the SSD \cite{liu2016ssd} by forcing it not to learn high-level abstraction and to keep low-level image features. 

Also, the depths of the earlier layers (eg. conv4\_3) used for small-sized object detection in SSD \cite{liu2016ssd} are very shallow. Therefore, in SSD, small objects can not be detected well because the representation power is insufficient to be used in the prediction as it is. To supplement this problem, we used a 3 $\times$ 3 convolution layer in branch2 of the Resblock as shown in Figure \ref{fig:Resblock} to reflect the peripheral contextual information.

Branch3 in the right side of Figure \ref{fig:Resblock} contains a deconvolution layer whose input is the feature maps of the consecutive layer. This is similar to a structure proposed in \cite{fu2017dssd} and \cite{Ren17CVPR}, and it is a proper method to propagate large contextual information to a small scale feature map so that even when detecting a small object, information about its surroundings is also utilized. This can reduce the cases of detecting a part of an actual object. Thus, it can be a remedy for the box-in-box problem described earlier. The effect of this is intuitively shown in the right side of Figure \ref{fig:boxinbox}. Finally, the proposed architecture in Figure \ref{fig:arch} can be expressed as follows:
\begin{multline} \label{eq:3.6}
 \mathbf{Detection} = \big\{ 
 \mathcal{P}_{1}(\mathbf{\hat{x}}_{s_{1},s_{2}}), 
 \mathcal{P}_{2}(\mathbf{\hat{x}}_{s_{2},s_{3}}), \\
 \dotsc,   
 \mathcal{P}_{k-1}(\mathbf{\hat{x}}_{s_{k-1},s_{k}}),
 \mathcal{P}_{k}(\mathbf{\hat{x}}_{s_{k}})
 \big\} ,
\end{multline}
where $
\mathbf{\hat{x}}_{a,b} = \mathcal{B}_{1}(\mathbf{x}_{a}) + \mathcal{B}_{2}(\mathbf{x}_{a}) + \mathcal{B}_{3}(\mathbf{x}_{b})
$ and 
$
\mathbf{\hat{x}}_{a} = \mathcal{B}_{1}(\mathbf{x}_{a}) + \mathcal{B}_{2}(\mathbf{x}_{a}).
$
Here, $\mathcal{B}_{1}$, $\mathcal{B}_{2}$ and $\mathcal{B}_{3}$ indicate branch1, branch2 and branch3, respectively.

\subsection{Unified Prediction Module}

\begin{figure}[t]
\begin{center}
\includegraphics[width=8cm]{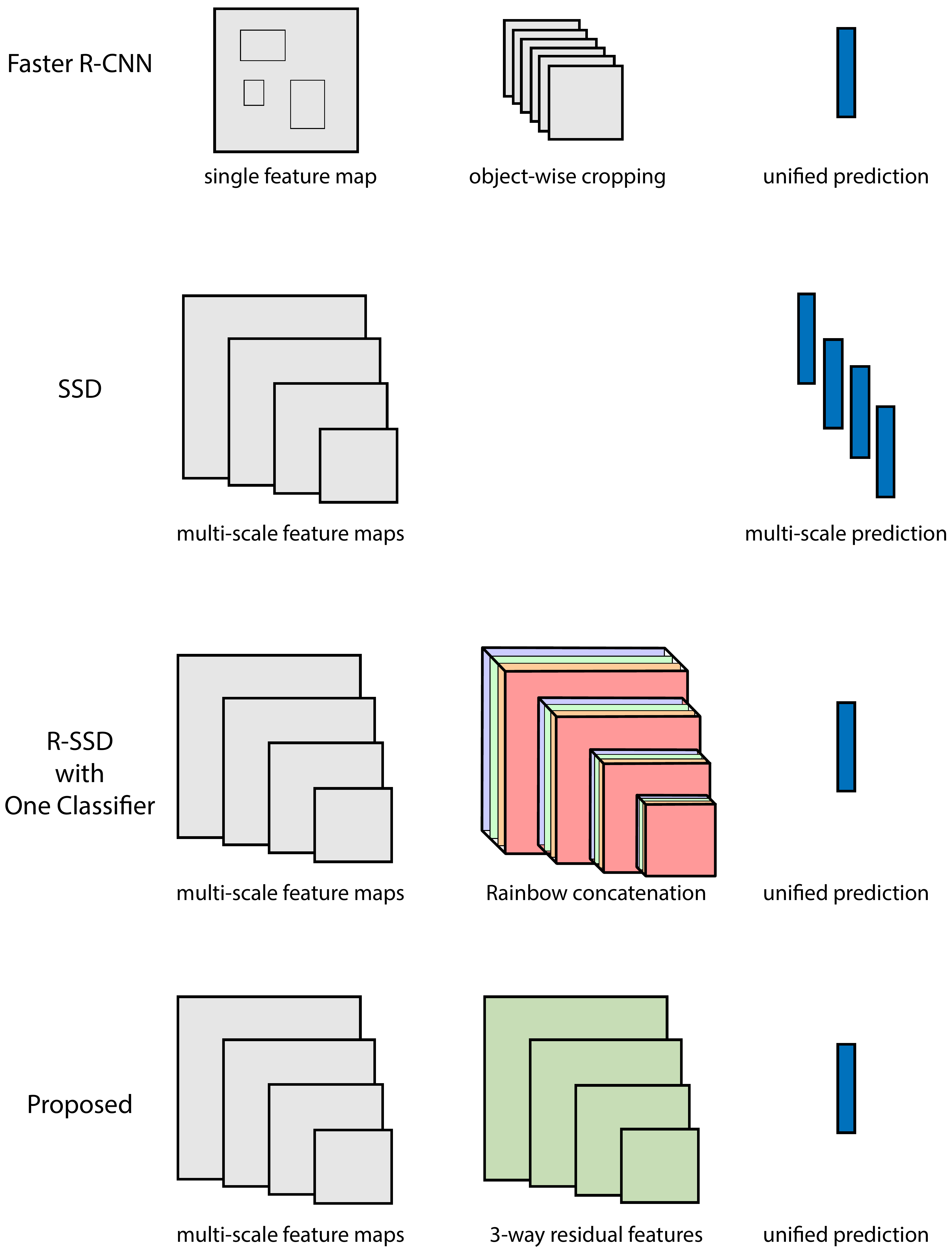}
\end{center}
\caption{
\textbf{Comparison of various object detection schemes}: a) R-CNN and its variants need object-wise cropping and the prediction is done by a common unified classifier. b) SSD does not need any cropping but requires a separate classifier for each scale of feature maps. c) R-SSD concatenates feature maps in different layers so that objects in each scale can be predicted with one unified classifier with the same amount of information. d) In the proposed method, Resblock takes the role of feature map concatenation and one unified classifier is used for prediction.  }
\label{fig:prediction}
\end{figure}

Detecting objects of various sizes has been recognized as an important problem in object detection. Traditionally, \cite{viola2004robust, dalal2005histograms, dollar2014fast} used a single classifier to predict multi-scale feature maps extracted from the image pyramid. There is another approach of using multiple classifiers on a single input image. The latter has the advantage of reducing the amount of computation for calculating feature maps. However, it requires an individual classifier for each object scale.

Since the neural network has been prevalent, the two-stage detectors applied RoI Pooling to the CNN output to extract feature maps of the same size from objects of different sizes. These feature maps were used as the input of a single classifier. Meanwhile, other methods using multi-scale features, such as SSD \cite{liu2016ssd}, adopted multiple classifiers since
feature maps in each scale differed not only in length but also in the underlying contextual information. 
In order to effectively learn the prediction layers of various scales, it is necessary to input objects of various scales. SSD could dramatically increase the detection performance through augmentation which transforms the size of input images.

R-SSD \cite{jeong2017enhancement} proposed the Rainbow concatenation which combines feature maps in different scales using pooling and deconvolution. This allows to set the depth of the input features for each prediction module to be the same. Thus, R-SSD could use a single classifier that shares the weight of multi-scale prediction modules. 
Similarly, the proposed 3-way Resblocks enforce all the feature maps to have the same depth of 256 as shown in Fig. \ref{fig:Resblock}. Thus, structurally, it is possible to unify convolution layers of different prediction modules 
like R-SSD. The idea of the unified prediction module is similar to \cite{jeong2017enhancement}, but our method is different from R-SSD in information contained in the input feature maps.

This approach makes differently-scaled feature maps have similar level of information. 
SSD \cite{liu2016ssd} used multiple features of various scales. This results in an improved performance of detecting small objects compared with YOLO \cite{redmon2016you,redmon2016yolo9000} which used only the last layer of the back-bone network.
However, since its earliest feature map is obtained from much shallower layers than the later feature maps, it still has a limitation of insufficient information for prediction.
Because unified prediction applies equally to feature maps of all scales, it forces the output of the 3-way Resblock between the feature map of the backbone and the prediction module to be learned at a similar information level.
It means that} unified prediction in combination with the residual feature block makes the feature maps in the earliest Resblocks rich in context. 
A brief summary of different object detection schemes is shown in Fig. \ref{fig:prediction}.

\section{Experiment} 
\label{sec:experiment}

We experimented the proposed method on PASCAL VOC 2007 \cite{everingham2010pascal}, PASCAL VOC 2012 and MS COCO datasets \cite{lin2014microsoft}.
Our implementation is based on the publicly available SSD \cite{liu2016ssd}\footnote{https://github.com/weiliu89/caffe/tree/ssd}. All of experiments results of SSD are the latest scores with data augmentation mentioned in \cite{fu2017dssd}.
For all the experiments, the reduced VGG-16 model \cite{simonyan2014very} pre-trained on the ILSVRC CLS-LOC dataset \cite{russakovsky2014imagenet} is used as the backbone network. For fair comparison, most of the settings are set to be the same as those of SSD except the number of proposals. It is different from SSD, because we used 6 default boxes in all the prediction layers for unified prediction while SSD used 4 for the conv4\_3 and the top layer, and 6 for the rest.

\begin{table}[t]
	\begin{center}
    \begin{tabular}{|l|c|}
    \hline
    Method & mAP \\
    \hline\hline
    SSD 300 & 77.5 \\
    \hline
    SSD 300 + 2WAY & 78.3\\ 
    SSD 300 + 2WAY + Unified Pred & 78.6 \\
    SSD 300 + 3WAY & 78.8 \\
    SSD 300 + 3WAY + Unified Pred & \textbf{79.2} \\
    \hline
    \end{tabular}
    \end{center}
    \caption{\textbf{PASCAL 2007} \texttt{test} detection results.}
    \label{tab:voc07}
\end{table}

\begin{table*}[pht]\small
	\begin{center}
	\setlength{\tabcolsep}{1.9pt}
      \begin{tabular*}{\textwidth}{l|c|c|c|cccccccccccccccccccc}
      \tiny Method & \tiny data & \tiny network & \tiny mAP & \tiny aero & \tiny bike & \tiny bird & \tiny boat & \tiny bottle & \tiny bus & \tiny car & \tiny cat & \tiny chair & \tiny cow & \tiny table & \tiny dog & \tiny horse & \tiny mbike & \tiny person & \tiny plant & \tiny sheep & \tiny sofa & \tiny train & \tiny tv \\
        \hline 
        \hline      
      \tiny SSD300~\cite{liu2016ssd} & \tiny 07++12& \tiny VGG &
      75.8 & 88.1 & 82.9 & 74.4 & 61.9 & 47.6 & 82.7 & 78.8 & 91.5 & 58.1 & 80.0 & 64.1 & 89.4 & 85.7 & 85.5 & 82.6 & 50.2 & 79.8 & 73.6 & 86.6 & 72.1\\
      \tiny SSD321~\cite{fu2017dssd} & \tiny 07++12 & \tiny Residual-101 & 75.4 & 87.9& 82.9&  73.7 & 61.5 & 45.3 & 81.4 & 75.6 & 92.6 & 57.4 & 78.3 & 65.0 & 90.8 & \textbf{86.8} & 85.8 & 81.5 & 50.3 &78.1 &75.3 &85.2 & 72.5 \\      
      \tiny DSSD321~\cite{fu2017dssd} & \tiny 07++12 & \tiny Residual-101&
      76.3 & 87.3 & 83.3 & 75.4 & \textbf{64.6} & 46.8 & 82.7 & 76.5 & \textbf{92.9} & 59.5 & 78.3 & 64.3 & \textbf{91.5} & 86.6 & \textbf{86.6} & 82.1 & \textbf{53.3} & 79.6 & \textbf{75.7} & 85.2 & 73.9 \\
      
      \tiny R-SSD300~\cite{jeong2017enhancement} & \tiny 07++12 & \tiny VGG &
      76.4 & 88.0 & 83.8 & 74.8 & 60.8 & 48.9 & \textbf{83.9} & 78.5 & 91.0 & 59.5 & 81.4 & 66.1 & 89.0 & 86.3 & 86.0 & 83.0 & 51.3 & 80.9 & 73.7 & 86.9 & 73.8 \\
      \tiny StairNet~\cite{woo2017stairnet} & \tiny 07++12 & \tiny VGG &
      76.4 & 87.7 & 83.1 & 74.6 & 64.2 & 51.3 & 83.6 & 78.0 & 92.0 & 58.9 & 81.8 & \textbf{66.2} & 89.6 & 86.0 & 84.9 & 82.6 & 50.9 & 80.5 & 71.8 & 86.2 & 73.5 \\
	\hline
      \tiny RUN2WAY300 & \tiny 07++12 & \tiny VGG &
      76.2 & \textbf{88.4} & 83.2 & 73.7 & 63.0 & 50.2 & 82.6 & 79.2 & 91.3 & 58.6 & 81.4 & 64.8 & 90.0 & 85.9 & 85.4 & 82.8 & 50.5 & 81.3 & 74.0 & 86.1 & 72.4 \\
      
      \tiny RUN3WAY300 & \tiny 07++12 & \tiny VGG &
      \textbf{77.1} & 88.2 & \textbf{84.4} & \textbf{76.2} & 63.8 & \textbf{53.1} & 82.9 & \textbf{79.5} & 90.9 & \textbf{60.7} & \textbf{82.5} & 64.1 & 89.6 & 86.5 & \textbf{86.6} & \textbf{83.3} & 51.5 & \textbf{83.0} & 74.0 & \textbf{87.6} & \textbf{74.4} \\
      \end{tabular*}
    	\end{center}
  		\caption{SSD300-based models on PASCAL 2012 \texttt{test}. Trained with
  \textbf{07++12} (07 \texttt{trainval} + 07 \texttt{test} + 12 \texttt{trainval}). 
  }
 \label{tab:voc12_300}
\end{table*}
\vspace{0.3cm}

\begin{table*}[t]\small
	\begin{center}
	\setlength{\tabcolsep}{1.9pt}
      \begin{tabular*}{\textwidth}{l|c|c|c|cccccccccccccccccccc}
      \tiny Method & \tiny data & \tiny network & \tiny mAP & \tiny aero & \tiny bike & \tiny bird & \tiny boat & \tiny bottle & \tiny bus & \tiny car & \tiny cat & \tiny chair & \tiny cow & \tiny table & \tiny dog & \tiny horse & \tiny mbike & \tiny person & \tiny plant & \tiny sheep & \tiny sofa & \tiny train & \tiny tv \\
        \hline 
        \hline      
        \tiny ION~\cite{bell2016inside} & \tiny 07+12+S & \tiny VGG & 76.4 & 87.5 & 84.7 & 76.8 & 63.8 & 58.3 & 82.6 & 79.0 & 90.9 & 57.8 & 82.0 & 64.7 & 88.9 & 86.5 & 84.7 & 82.3 & 51.4 & 78.2 & 69.2 & 85.2 & 73.5 \\
       
       \tiny Faster~\cite{he2016deep} & \tiny 07++12 & \tiny Residual-101 & 73.8 & 86.5 & 81.6 & 77.2 & 58.0 & 51.0 & 78.6 & 76.6 & 93.2 & 48.6 & 80.4 & 59.0 & 92.1 & 85.3 & 84.8 & 80.7 & 48.1 & 77.3 & 66.5 & 84.7 & 65.6 \\ 
              
       \tiny R-FCN~\cite{li2016r} & \tiny 07++12& \tiny Residual-101 & 77.6 & 86.9 & 83.4 & \textbf{81.5}& 63.8& \textbf{62.4} & 81.6 & 81.1 & 93.1 & 58.0 & 83.8 & 60.8& 92.7 & 86.0 & 84.6 & 84.4 & \textbf{59.0} & 80.8 & 68.6& 86.1 & 72.9 \\
       \hline
      \tiny SSD512~\cite{liu2016ssd} & \tiny 07++12&
     \tiny VGG & 78.5 & 90.0 & 85.3 & 77.7 & 64.3 & 58.5 & 85.1 & 84.3 & 92.6 & 61.3 & 83.4 & 65.1 & 89.9 & 88.5 & \textbf{88.2} & 85.5 & 54.4 & 82.4 & 70.7 & 87.1 & 75.6\\
     \tiny SSD 513~\cite{fu2017dssd} & \tiny 07++12 & 
     \tiny Residual-101 & 79.4 & 90.7 & \textbf{87.3} & 78.3 & 66.3 & 56.5 & 84.1 & 83.7 &  94.2 & 62.9 & 84.5 & 66.3 & 92.9 & 88.6 & 87.9 & 85.7 & 55.1 & 83.6  & 74.3 & \textbf{88.2} & \textbf{76.8} \\

     \tiny DSSD 513~\cite{fu2017dssd} & \tiny 07++12& 
     \tiny Residual-101 & \textbf{80.0} & \textbf{92.1} & 86.6 & 80.3 & \textbf{68.7} & 58.2 & 84.3 & 85.0 & \textbf{94.6} & \textbf{63.3} & \textbf{85.9} & 65.6 & \textbf{93.0} & 88.5 & 87.8 & 86.4 & 57.4 & \textbf{85.2} & 73.4 & 87.8 & \textbf{76.8}  \\
     \hline
     \tiny RUN2WAY512 & \tiny 07++12& 
     \tiny VGG & 79.3 & 89.7 & 87.1 & 79.2 & 65.6 & 61.3 & \textbf{85.3} & 85.0 & 92.9 & 60.6 & 83.8 & \textbf{66.4} & 90.6 & 88.6 & 88.1 & 86.1 & 54.8 & 84.6 & 72.5 & 87.4 & 75.8  \\
     
     \tiny RUN3WAY512 & \tiny 07++12& 
     \tiny VGG & 79.8 & 90.0 & \textbf{87.3} & 80.2 & 67.4 & \textbf{62.4} & 84.9 & \textbf{85.6} & 92.9 & 61.8 & 84.9 & 66.2 & 90.9 & \textbf{89.1} & 88.0 & \textbf{86.5} & 55.4 & 85.0 & 72.6 & 87.7 & \textbf{76.8}  \\

		\end{tabular*}
    	\end{center}
  		\caption{SSD500-based models and other two-stage detectors on PASCAL 2012 \texttt{test}. Trained with \textbf{07++12} (07 \texttt{trainval} + 07 \texttt{test} + 12 \texttt{trainval}).
}
\label{tab:voc12}
\end{table*}

\begin{table*}[tpb]
	\centering
	\setlength{\tabcolsep}{2.6pt}
	\begin{tabular*}{\textwidth}{l|c|c|C{3.0em}C{2.2em}C{2.2em}|C{2em}C{2em}C{2em}|C{2.1em}C{2.1em}C{2.1em}|C{2em}C{2em}C{2em}}
     
    	\multirow{2}{*}{Method} & \multirow{2}{*}{data} &     
\multirow{2}{*}{network} &        
\multicolumn{3}{c|}{\scriptsize{Avg. Precision, IoU:}} & \multicolumn{3}{c|}{\scriptsize{Avg. Precision, Area:}} & \multicolumn{3}{c|}{\scriptsize{Avg. Recall, \#Dets:}} & \multicolumn{3}{c}{\scriptsize{Avg. Recall, Area:}}\\
        & & &\small{0.5:0.95} & 0.5 & 0.75 & S & M & L & 1 & 10 & 100 & S & M & L\\
        \hline
     
        Faster~\cite{ren2015faster} & trainval & VGG & 21.9 & 42.7 & - & - & - & - & - & - & - & - & - & -\\
        ION~\cite{bell2016inside} & train & VGG & 23.6 & 43.2 & 23.6 & 6.4 & 24.1 & 38.3 & 23.2 & 32.7 & 33.5 & 10.1 & 37.7 & 53.6\\

          R-FCN~\cite{li2016r} & trainval & Residual-101 & 29.9&51.9  & - & 10.8 & 32.8 & 45.0 & - & - &-&-&-&-  \\
          RetinaNet~\cite{lin2017focal} & trainval & Residual-101
 & \textbf{39.1} & \textbf{59.1}  & \textbf{42.3} & \textbf{21.8} & \textbf{42.7} & \textbf{50.2} & - & - &-&-&-&-  \\
        \hline
        SSD300~\cite{liu2016ssd} & trainval35k & VGG & 25.1 & 43.1 & 25.8 & 6.6 & 25.9 & 41.4 & 23.7 & 35.1 & 37.2 & 11.2 & 40.4 & 58.4\\
        SSD321~\cite{fu2017dssd} & trainval35k & Residual-101 & 28.0 & 45.4 & 29.3 & 6.2 & 28.3 & 49.3  & 25.9 & 37.8 & 39.9 & 11.5 & 43.3 & 64.9  \\
        DSSD321~\cite{fu2017dssd} & trainval35k & Residual-101 & 
        28.0 & 46.1 & 29.2 & 7.4 & 28.1 & 47.6 &  25.5 & 37.1 & 39.4& 12.7 & 42.0 & 62.6\\
        RUN2WAY300 & trainval35k & VGG & 27.4 & 46.1 & 28.4 & 8.9 & 27.9 & 43.8 & 25.0 & 37.3 & 39.5 & 14.6 & 42.6 & 59.8\\
        RUN3WAY300 & trainval35k & VGG & 28.0 & 47.5 & 28.9 & 9.9 & 28.6 & 43.9 & 25.3 & 38.0 & 40.5 & 16.2 & 43.8 & 60.2\\
        
        \hline 
        SSD512~\cite{liu2016ssd} & trainval35k & VGG & 28.8 & 48.5 & 30.3 & 10.9 & 31.8 & 43.5 & 26.1 & 39.5 & 42.0 & 16.5 & 46.6 & 60.8\\      
        SSD513~\cite{fu2017dssd} & trainval35k & Residual-101 & 31.2 & 50.4 & 33.3 & 10.2 & 34.5 & 49.8 & 28.3 & 42.1 & 44.4 & 17.6 & \textbf{49.2} & 65.8 \\
        DSSD513~\cite{fu2017dssd} & trainval35k & Residual-101 & 33.2 & 53.3 & 35.2 & 13.0 & 35.4 & 51.1 & \textbf{28.9} & \textbf{43.5} & \textbf{46.2} & 21.8 & 49.1 & \textbf{66.4} \\ 
        RUN2WAY512 & trainval35k & VGG & 31.7 & 52.1 & 33.6 & 13.2 & 33.9 & 46.5 & 27.7 & 42.2 & 44.7 & 22.0 & 47.9 & 62.7\\
        RUN3WAY512 & trainval35k & VGG & 32.4 & 53.5 & 34.2 & 14.7 & 34.0 & 46.7 & 28.0 & 43.0 & 45.8 & \textbf{24.4} & 48.1 & 63.4\\
        
    \end{tabular*}
    \vspace{0.2cm}
    \caption{\textbf{MS COCO \texttt{test-dev} detection results.}}
    \label{tab:coco}
\end{table*}

\paragraph{Ablation Study on PASCAL VOC2007}
We trained our model on  VOC2007 \texttt{trainval} and VOC2012 \texttt{trainval}. 
We set the batch size as 32. For the training of the 2-way model, we used learning rate of $10^{-3}$ initially, then it decreased by a factor of 10 at 80k and 100k iterations, respectively. The training was terminated at 120k iterations. For the 3-way model, we froze all the weights of the pre-trained 2-way model except the prediction module, then fine-tuned the network using the learning rate of $10^{-3}$ for 40k iterations, $10^{-4}$ for the next 20k iterations, and $10^{-5}$ for the final 10k iterations. The end-to-end training was also applied on the 3-way model, but the results were worse than the above training strategy.

Table \ref{tab:voc07} shows our result on PASCAL VOC 2007 \texttt{test} set. 
Here, \textit{Unified Pred} is the proposed unified prediction module and the prediction modules for the ones without this indication were trained separately as in the original SSD. 
As mentioned above, each 3-way model was fine-tuned on the corresponding 2-way model. In this experiment, we observed that the proposed model with only 2-way Resblock without the deconvolution path achieved 1.1\% higher mAP than that of SSD. The 3-way model which further utilizes deconvolution layers was up to 0.6\% higher than the 2-way model. The unified prediction  module made better advance in the 3-way model than the 2-way model, which scored 79.2\% and 78.4\% respectively.

\paragraph{PASCAL VOC 2012}
For VOC 2012 \texttt{test}, we trained models on 07++12 dataset consisting 07\texttt{trainval}, 07\texttt{test} and 12\texttt{trainval}. First, we performed an experiment applying the 2-way Resblock in combination with the unified prediction, then, another experiment was performed using the 3-way Resblock with unified prediction after freezing the weights of the contained 2-way Resblock.
 
 Table \ref{tab:voc12_300} shows the VOC 2012 \texttt{test} results 
 of RUN300 and other models based on SSD300 \cite{liu2016ssd}.
 The proposed model, RUN300, has a big performance improvement compared to the base model SSD300. Especially, the 3-way model achieved 77.1\% mAP, outperforming other SSD-based models. In addition, it showed improvement of 0.7\% mAP compared to StairNet \cite{woo2017stairnet} which uses FPN \cite{lin2016feature} and unified prediction. From this result, we can conjecture that 
the proposed 3-way Resblock is more effective than FPN.
 
Table \ref{tab:voc12} shows results of RUN512 models and others. The 3-way model achieved 79.8\% mAPs, which is 1.3\% better than that of SSD512 \cite{liu2016ssd}. It performs slightly worse than DSSD513 \cite{fu2017dssd}, which is probably because the ResNet-101 \cite{he2016deep} backbone of DSSD513 produces better features for larger input images than VGG-16 \cite{simonyan2014very} of SSD and ours.

\paragraph{COCO}
For fair comparison with SSD \cite{liu2016ssd}, most of the hyper-parameters required for training were set to the same as SSD.
For training 2-way models, we used a learning rate of $10^{-3}$ for the first 240k iterations, $10^{-4}$ for the next 120k iterations and $10^{-5}$ for the last 40k. For training 3-way models, we used a learning rate of $10^{-3}$ for the first 120k iterations, $10^{-4}$ for the next 60k iterations and $10^{-5}$ for the last 20k, which are exactly half of those for the 2-way models. Other parameters such as scales and aspect ratios of the prior box were identical to those of SSD.

\begin{table}[t]
\begin{center}
	\centering
    \setlength{\tabcolsep}{3pt}
    \begin{tabular}{l|c|c|c|c}

	Method & network & mAP & FPS & GPU \\
    \hline
        Faster R-CNN \cite{ren2015faster} &VGG16 &73.2 & 7 & \scriptsize{Titan X} \\
        Faster R-CNN \cite{he2016deep} & Residual-101 & 76.4 & 2.4& \scriptsize{K40} \\
		R-FCN \cite{li2016r}& Residual-101 &   80.5 & 9 & \scriptsize{Titan X} \\

        \hline
        SSD300~\cite{liu2016ssd} & VGG16 & 77.5 & 54.5*& \scriptsize{Titan X} \\
        SSD321~\cite{fu2017dssd} & Residual-101 &77.1 &16.4 & \scriptsize{Titan X} \\ 
        DSSD321~\cite{fu2017dssd} & Residual-101 &78.6 &11.8 & \scriptsize{Titan X} \\ 
        R-SSD300 \cite{jeong2017enhancement} & VGG16 & 78.5 & 37.1* & \scriptsize{Titan X} \\
        StairNet \cite{woo2017stairnet} & VGG16 & 78.8 & 30 & \scriptsize{Titan X Pascal} \\
        \hdashline[.4pt/1pt]
        \multirow{2}{*}{RUN2WAY300} & \multirow{2}{*}{VGG16} & \multirow{ 2}{*}{ 78.6 } & 41.8 & \scriptsize{Titan X} \\
        &&&58.4 & \scriptsize{Titan X Pascal} \\
        \hdashline[.4pt/1pt]
        \multirow{2}{*}{RUN3WAY300} & \multirow{2}{*}{VGG16} & \multirow{2}{*}{79.2} & 40.0 & \scriptsize{Titan X} \\
        &&&56.3 & \scriptsize{Titan X Pascal} \\
        
        \hline 
        SSD512 \cite{liu2016ssd} & VGG16 & 79.5 & 24.5*& \scriptsize{Titan X} \\
        SSD513 \cite{fu2017dssd} & Residual-101 &80.6 &8.0 & \scriptsize{Titan X}  \\
        DSSD513 \cite{fu2017dssd} & Residual-101 &81.5 &6.4 & \scriptsize{Titan X}  \\
        R-SSD512 \cite{jeong2017enhancement} & VGG16 & 80.8 & 15.8* & \scriptsize{Titan X} \\
        \hdashline[.4pt/1pt]
        \multirow{2}{*}{RUN2WAY512} & \multirow{2}{*}{VGG16} & \multirow{2}{*}{80.6} & 20.1 & \scriptsize{Titan X} \\
        &&&31.8 & \scriptsize{Titan X Pascal} \\
        \hdashline[.4pt/1pt]
        \multirow{2}{*}{RUN3WAY512} & \multirow{2}{*}{VGG16} & \multirow{2}{*}{80.9} & 19.5 & \scriptsize{Titan X} \\
        &&&29.8 & \scriptsize{Titan X Pascal} \\
	\end{tabular}
    \end{center}
    \caption{\textbf{Speed \& Accuracy on PASCAL VOC2007 \texttt{test}.} * is measured by ourselves.}
    \label{tab:voc07_with_fps}
\end{table}

Table \ref{tab:coco} shows the performance of various methods on MS COCO \texttt{test-dev}.
Despite the proposed methods use a relatively shallow network, VGG-16 \cite{simonyan2014very}, they achieved enough performance to compare with other methods which use a very deep network. The fourth column indicates that RUN3WAY300 achieved 2.9\% better mAP compared to SSD300 \cite{liu2016ssd}. It was the same performance with SSD321 and DSSD321~\cite{fu2017dssd}, which adopted ResNet-101~\cite{he2016deep} as their back-bone network. Also, RUN3WAY512 achieved 3.6\% better mAP than SSD512. In particular, RUN3WAY512 achieved the highest average precision and recall for small objects among compared methods except RetinaNet. It means that the proposed Resblock is a quite effective module to enhance low-level feature maps.

\paragraph{Speed vs Accuracy}

\begin{figure}[t]
\begin{center}
\includegraphics[height=6.5cm]{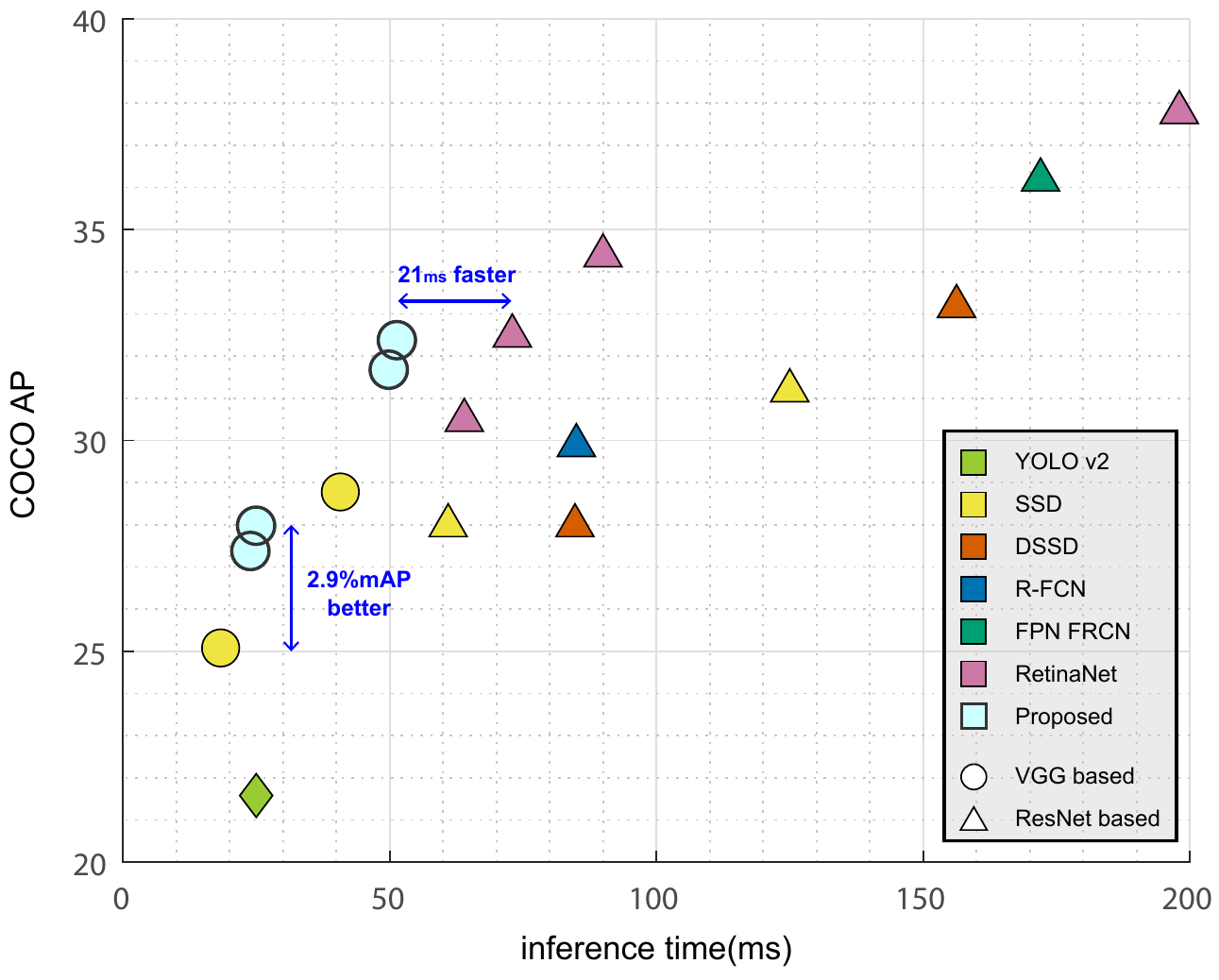}
\end{center}
\caption{
\textbf{Speed vs. Accuracy of recent methods using public numbers on COCO}. Our results (sky blue circles) are measured on Titan X. (Best viewed in color.)}
\label{fig:speed_vs_speed_on_coco}
\end{figure}

\begin{figure*}[h]
    \begin{subfigure}[b]{\textwidth}
    \centering
	\includegraphics[width=0.19\linewidth]{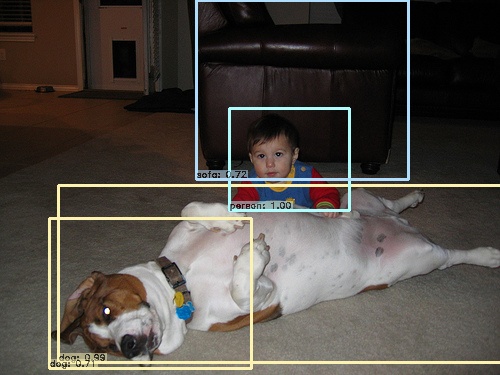}\hfill
    \includegraphics[width=0.19\linewidth]{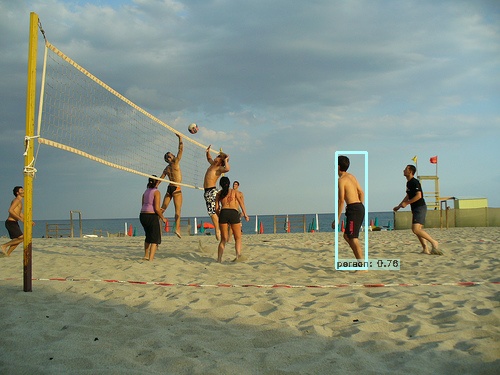}\hfill
    \includegraphics[width=0.19\linewidth]{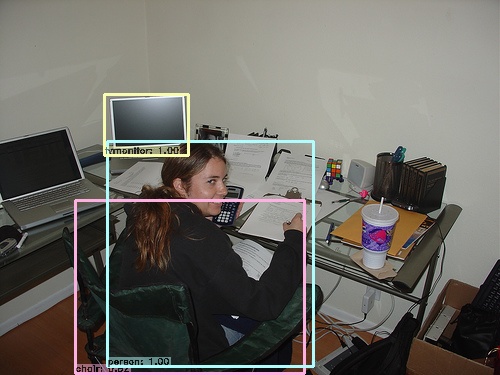}\hfill
    \includegraphics[width=0.19\linewidth]{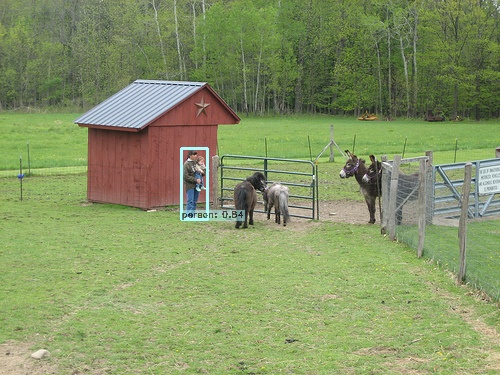}\hfill
    \includegraphics[width=0.19\linewidth]{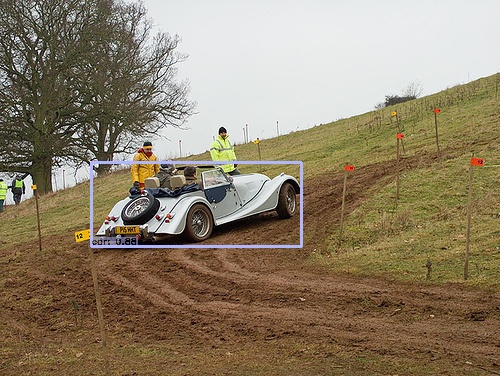}
\\
 \vspace{0.1cm}
    \includegraphics[width=0.19\linewidth]{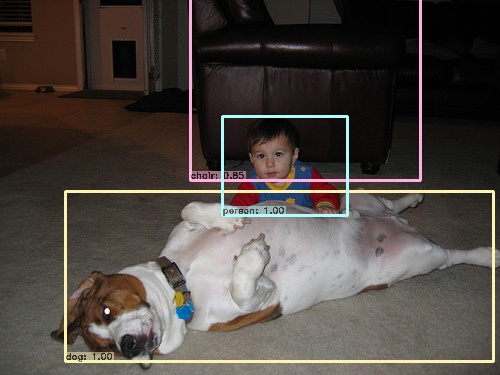}\hfill
    \includegraphics[width=0.19\linewidth]{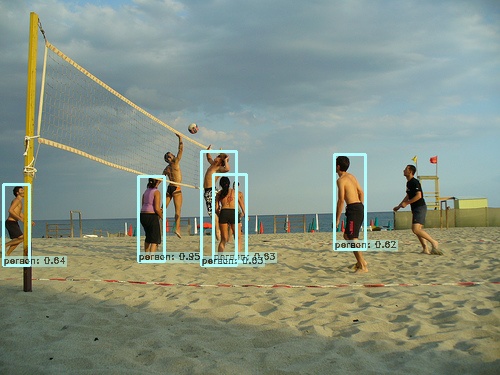}\hfill
    \includegraphics[width=0.19\linewidth]{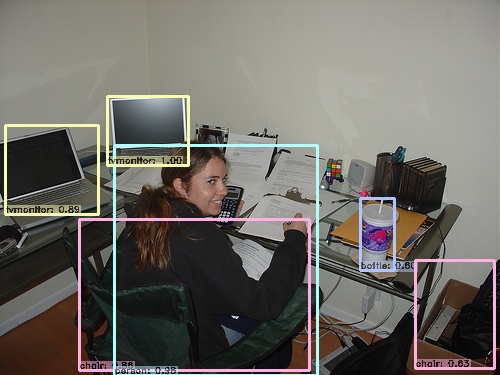}\hfill
    \includegraphics[width=0.19\linewidth]{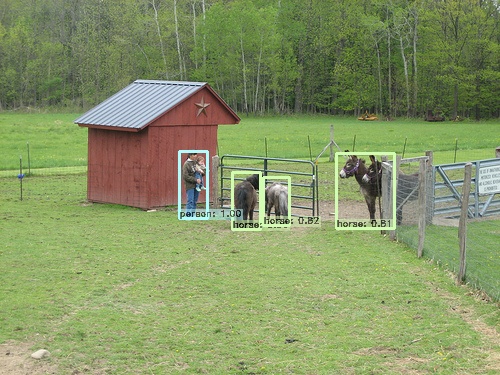}\hfill
    \includegraphics[width=0.19\linewidth]{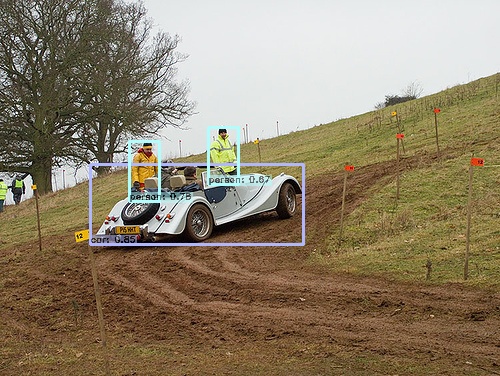}
\\
  \label{sfig:results}
  \end{subfigure}
  \caption{
    \textbf{Detection examples of RUN300 3-way on PASCAL VOC 2012 test set compared with SSD300 model.} For each pair, the up side is the result of SSD and down side is the result of RUN. We show detections with scores higher than 0.6. Each color corresponds to an object category. }
\end{figure*}

\begin{figure*}[ht]
    \begin{subfigure}[b]{\textwidth}
    \centering  
    \includegraphics[height=0.13\linewidth]{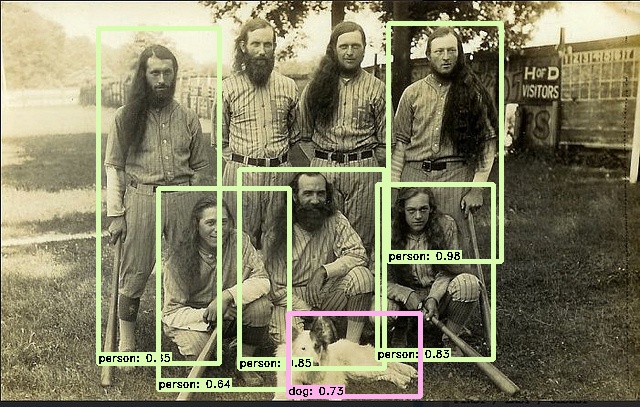}\hfill
    \includegraphics[height=0.13\linewidth]{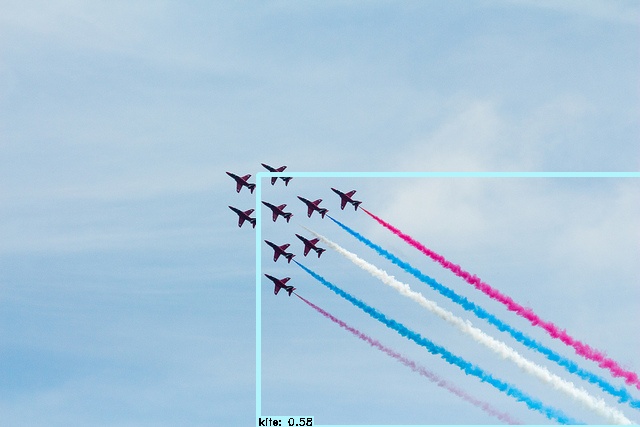}\hfill
    \includegraphics[height=0.13\linewidth]{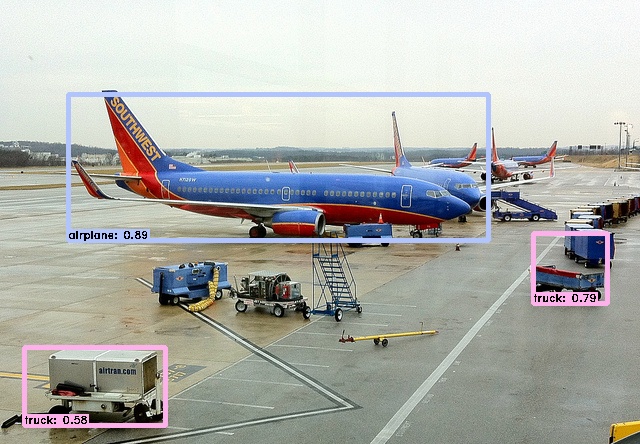}\hfill
    \includegraphics[height=0.13\linewidth]{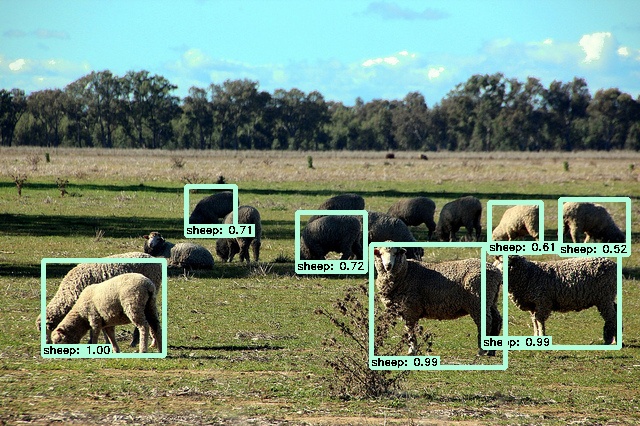}\hfill
    \includegraphics[height=0.13\linewidth]{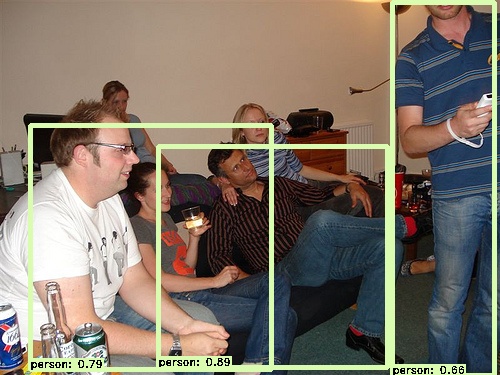}
\\
 \vspace{0.1cm}
    \includegraphics[height=0.13\linewidth]{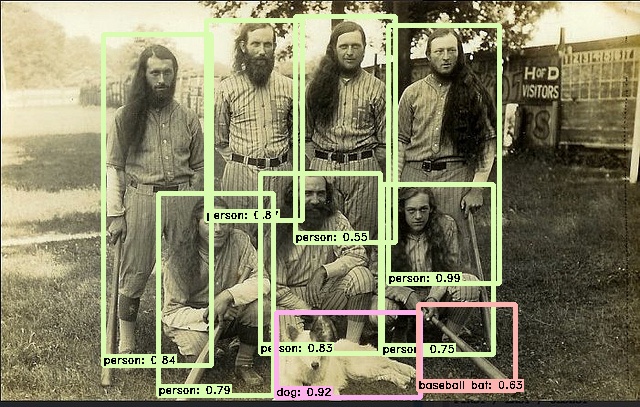}\hfill
    \includegraphics[height=0.13\linewidth]{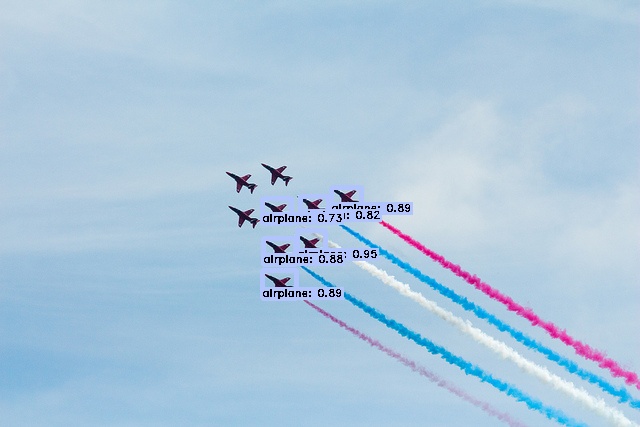}\hfill
    \includegraphics[height=0.13\linewidth]{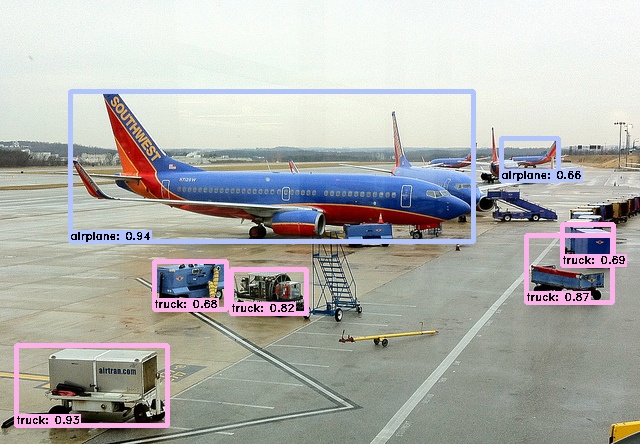}\hfill
    \includegraphics[height=0.13\linewidth]{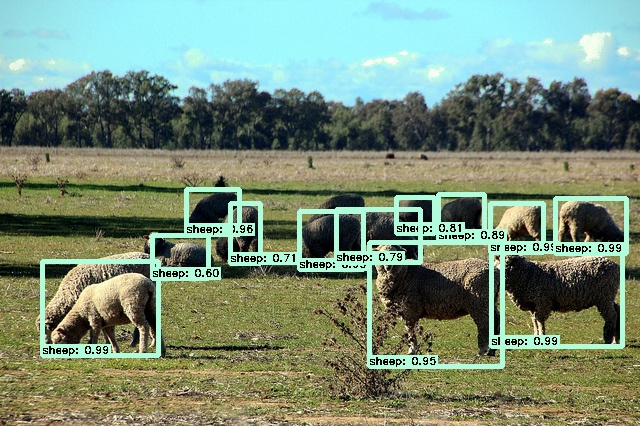}\hfill
    \includegraphics[height=0.13\linewidth]{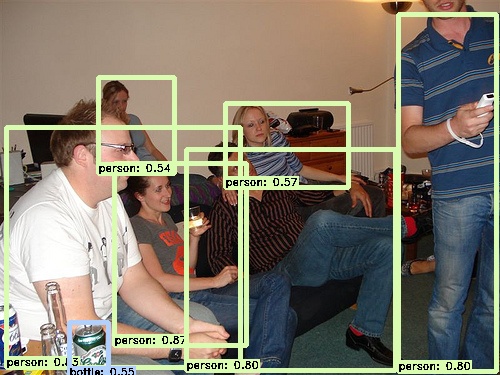}
\\
  \label{sfig:results_coco}
  \end{subfigure}
  \caption{
    \textbf{Detection examples of RUN300 3-way on MS COCO \texttt{test-dev} set compared with SSD300 model.} For each pair, the up side is the result of SSD and down side is the result of RUN. We show detections with scores higher than 0.5. Each color corresponds to an object category. }
\end{figure*}

The single stage detectors, which are represented by YOLO \cite{redmon2016you} and SSD \cite{liu2016ssd}, proposed end-to-end neural networks that removed the RoI Pooling of two-stage detectors. They have achieved a lot of speed improvements, but they could not avoid the loss of accuracy. Conversely, recent single stage detectors have been studied to improve performance, while suffering the loss of speed. Unlike other approaches, the proposed RUN is designed to maximize performance at high speeds on the VGG-16 \cite{simonyan2014very} backbone, which has significantly fewer layers and parameters than ResNet \cite{he2016deep}. The experimented results demonstrate the performance improvement of RUN.

Table\ref{tab:voc07_with_fps} shows that our method outperforms other competitors with less loss of speed. Our experiments were tested using Titan X GPU, cuDNN v5.1 and Intel I7-6700@3.4GHz. For exact comparison, we measured FPS of some methods on the same environment and marked * in the table.

In Figure \ref{fig:speed_vs_speed_on_coco}, we show the trade-off relation between the detection accuracy and inference time by plotting the results of RUN and other methods on COCO \texttt{test-dev}. The RUN-3way-300 model (25.0ms, 28.0\% mAP) is 36\% slower but 2.9\% better in mAP than the SSD300~\cite{liu2016ssd} model (18.3ms, 25.1\% mAP). It is about 60\% faster than ResNet-101 based SSD321~\cite{fu2017dssd} model (61ms, 28.0\% mAP) that has a similar performance. Likewise, the RUN-3way-512 (51.4ms, 32.4\% mAP) is 26\% slower but 3.6\% better in mAP than the SSD512 model (40.8ms, 28.8\% mAP). It is about 44\% faster than RetinaNet-50-500~\cite{lin2017focal}  (73ms, 32.5\% mAP) that has a similar performance.

In addition, we measured FPS of our methods on Titan X Pascal with the other environment kept the same. Table~\ref{tab:voc07_with_fps} shows that even the most complex version of our method, RUN3WAY512, can works in real time (29.8 FPS) on Tital X Pascal.

\section{Conclusion}
\label{sec:conclusion}
The proposed RUN architecture for object detection was originated from the awareness of the contradictory requirements for multi-scale features that they should contain low-level information on an image as well as high-level information on objectness. The proposed 3-way Resblock alleviated the gradient exploitation problem and enriched contextual information, an important element of prediction. We also showed that the generalization performance of multi-scale prediction can be improved by integrating the separate prediction modules into one unified prediction module. This approach, which can be seen to be somewhat simple, resulted in outstanding performance on the PASCAL VOC test. The results on COCO dataset also show how fast and efficient our algorithms are. We expect the proposed method be not restricted to SSD-based methods but also applicable to other structures utilizing multi-scale features.

{\small
\bibliographystyle{ieee}
\bibliography{egbib}
}

\clearpage

\begin{figure*}[ht]
\begin{center}
	\begin{subfigure}{\textwidth}    	
	    \includegraphics[height=0.177\linewidth]{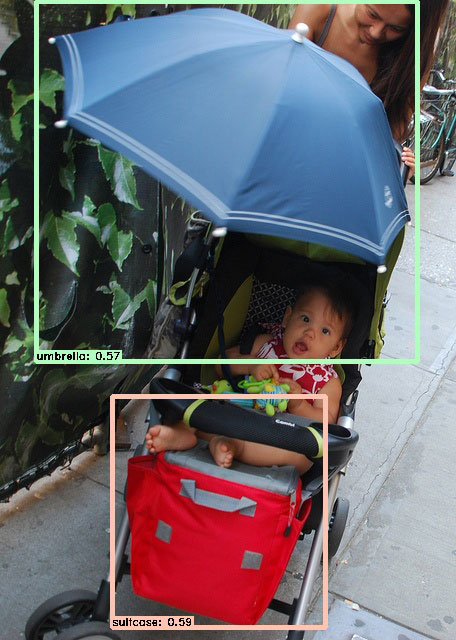}
        \hfill
        \includegraphics[height=0.177\linewidth]
        {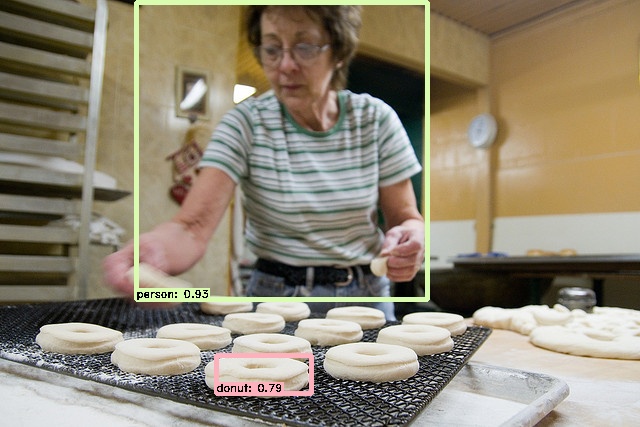}
        \hfill
        \includegraphics[height=0.177\linewidth]{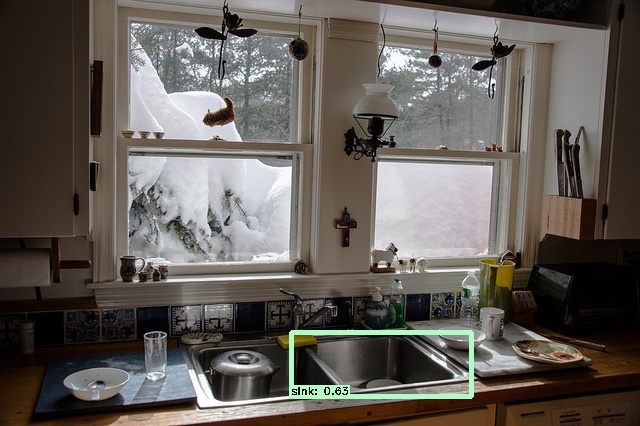}
\hfill
        \includegraphics[height=0.177\linewidth]{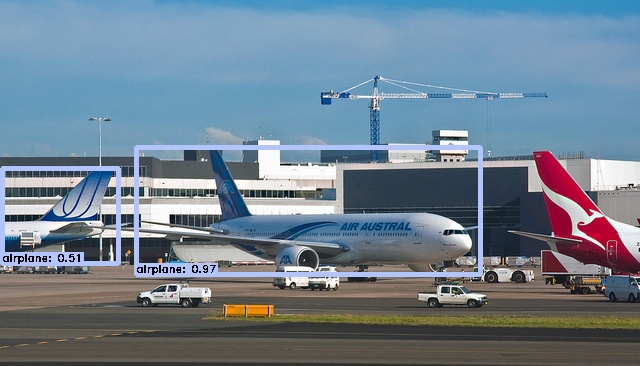} \\
        \includegraphics[height=0.177\linewidth]{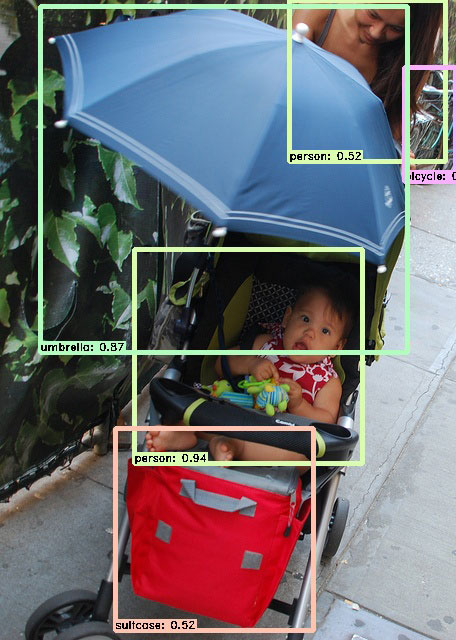}
        \hfill
        \includegraphics[height=0.177\linewidth]{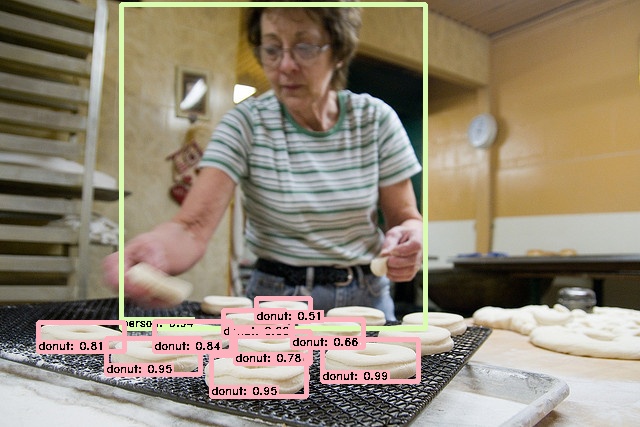}
        \hfill
        \includegraphics[height=0.177\linewidth]{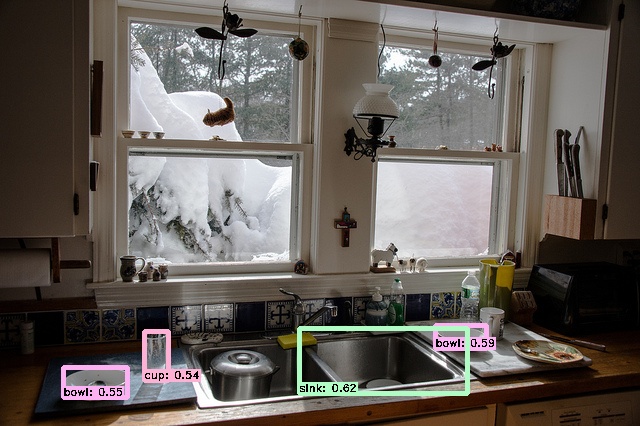}
        \hfill
        \includegraphics[height=0.177\linewidth]{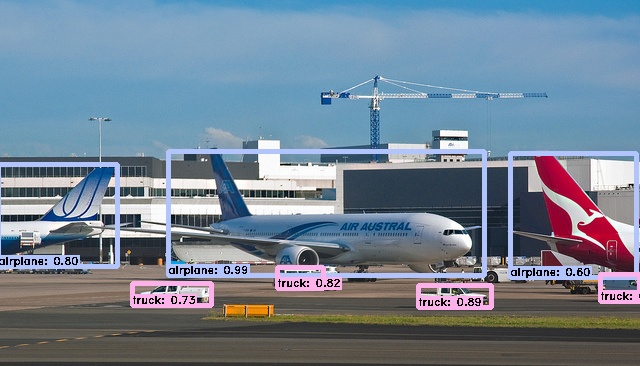}
        \\
        
        \includegraphics[height=0.174\linewidth]{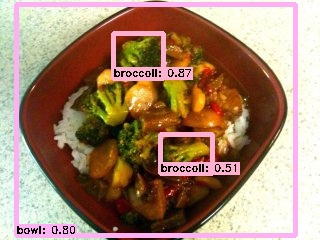}
        \hfill
	    \includegraphics[height=0.174\linewidth]{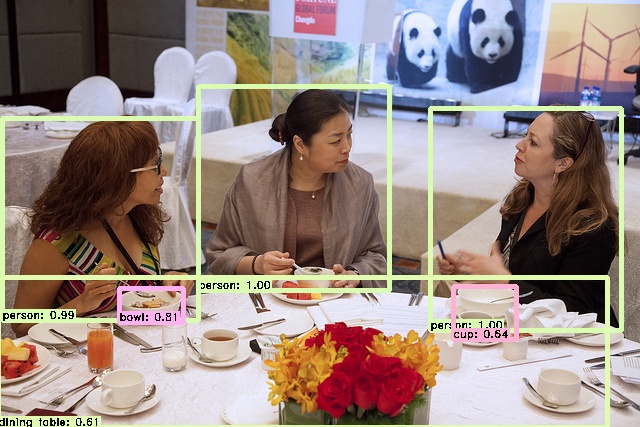}
        \hfill
        \includegraphics[height=0.174\linewidth]{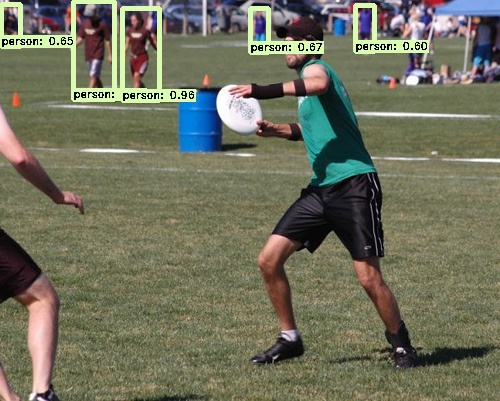}
        \hfill
        \includegraphics[height=0.174\linewidth]{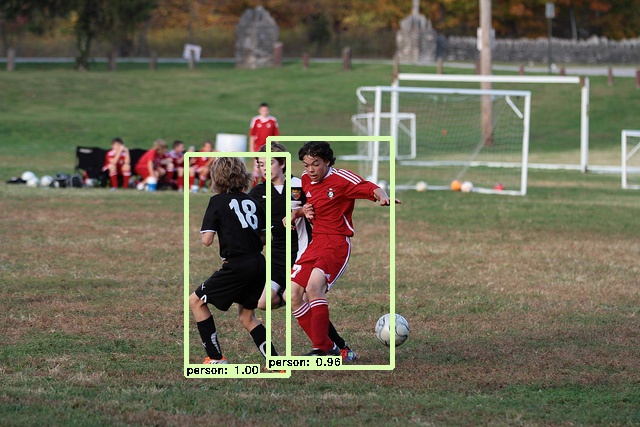}\\
        \includegraphics[height=0.174\linewidth]{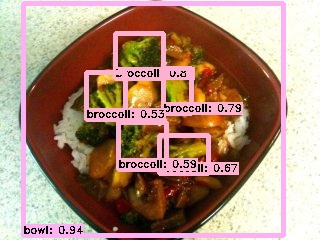}
        \hfill
        \includegraphics[height=0.174\linewidth]{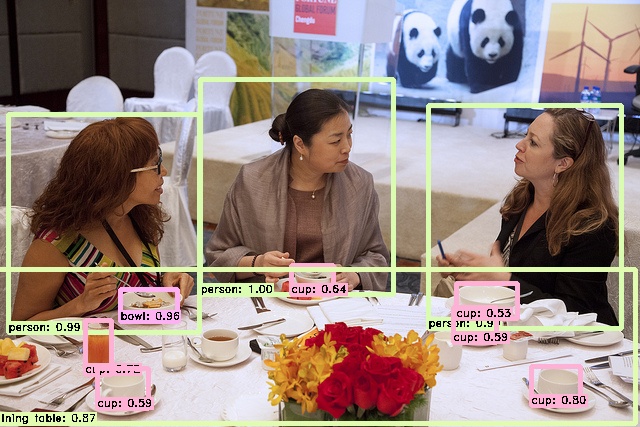}
        \hfill
        \includegraphics[height=0.174\linewidth]{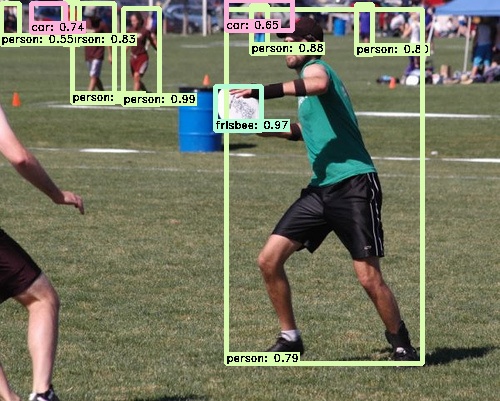}
        \hfill
        \includegraphics[height=0.174\linewidth]{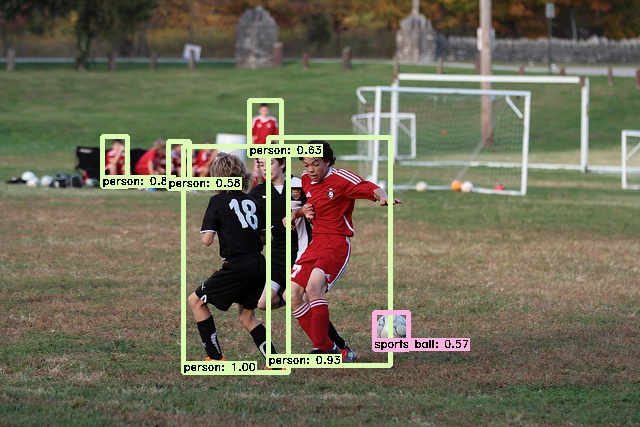}
        \\
        
        \includegraphics[height=0.173\linewidth]{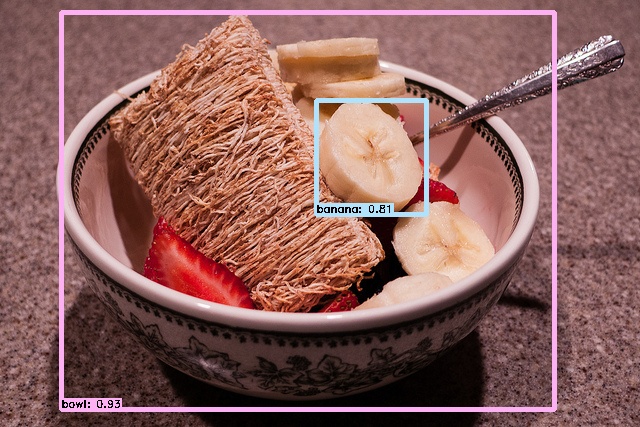}
        \hfill
        \includegraphics[height=0.173\linewidth]{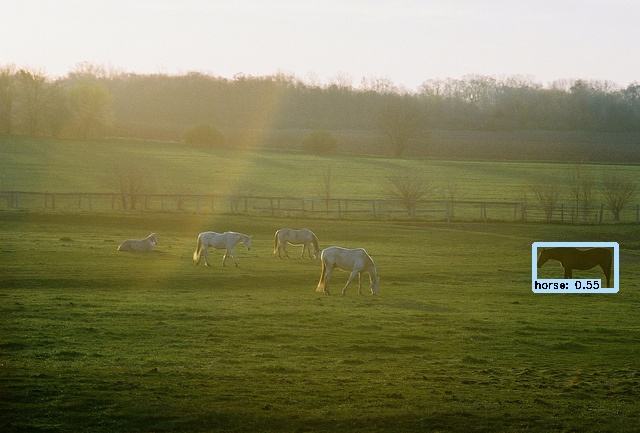}
        \hfill
        \includegraphics[height=0.173\linewidth]{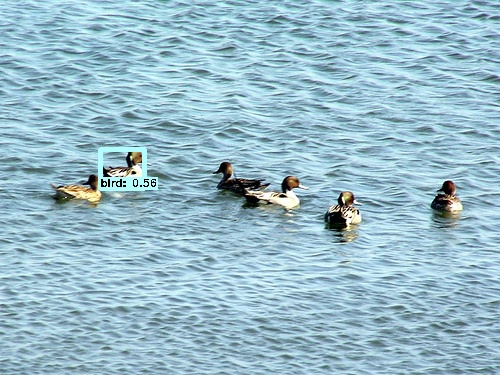}
        \hfill
        \includegraphics[height=0.173\linewidth]{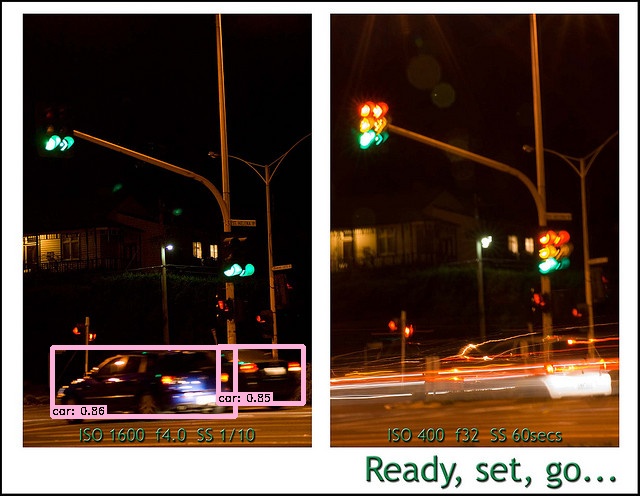}\\
        \includegraphics[height=0.173\linewidth]{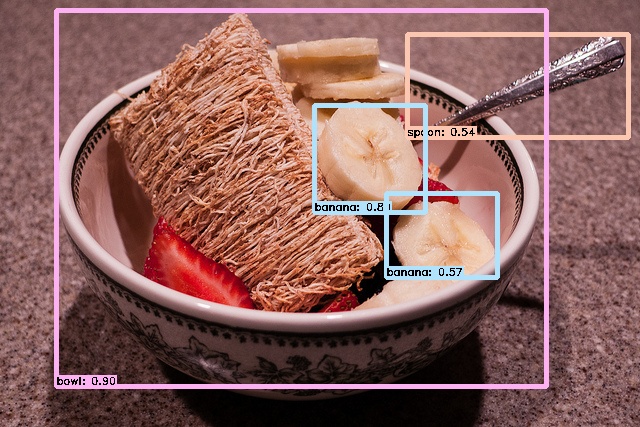}
        \hfill
        \includegraphics[height=0.173\linewidth]{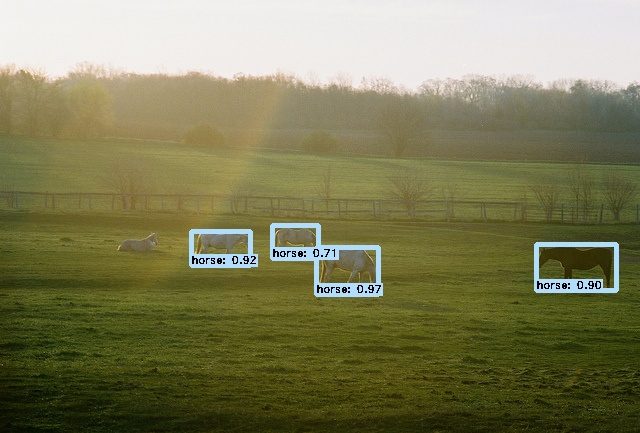}
        \hfill
        \includegraphics[height=0.173\linewidth]{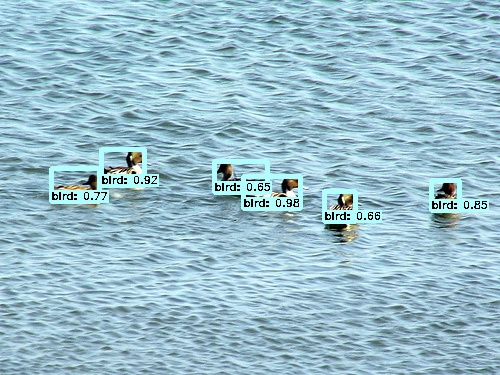}
        \hfill
        \includegraphics[height=0.173\linewidth]{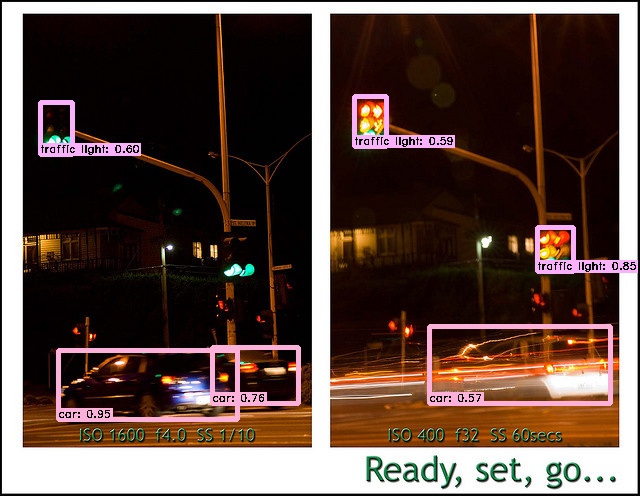}        \\
        \end{subfigure}
\end{center}
\caption{
\textbf{Detection examples of RUN300 3-way on MS COCO \texttt{test-dev} set compared with SSD300 model.} For each pair, the up side is the result of SSD and down side is the result of RUN. We show detections with scores higher than 0.5. Each color corresponds to an object category. Our method is especially good for detecting small objects compared to SSD.}
\label{fig:add_figures1}
\end{figure*}

\clearpage
\begin{figure*}[ht]
\begin{center}
	\begin{subfigure}{\textwidth}    	
	    \includegraphics[height=0.255\linewidth]{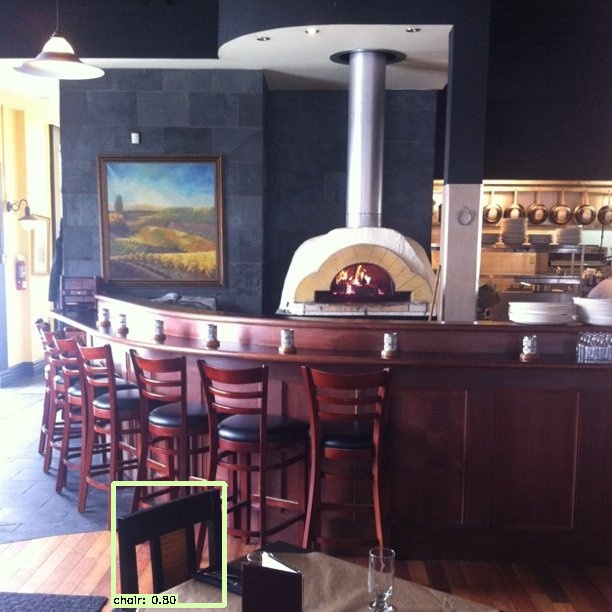}       		\hfill
        \includegraphics[height=0.255\linewidth]{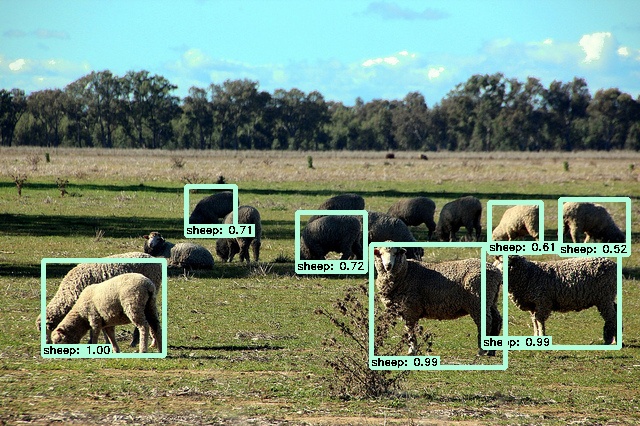}
\hfill
        \includegraphics[height=0.255\linewidth]{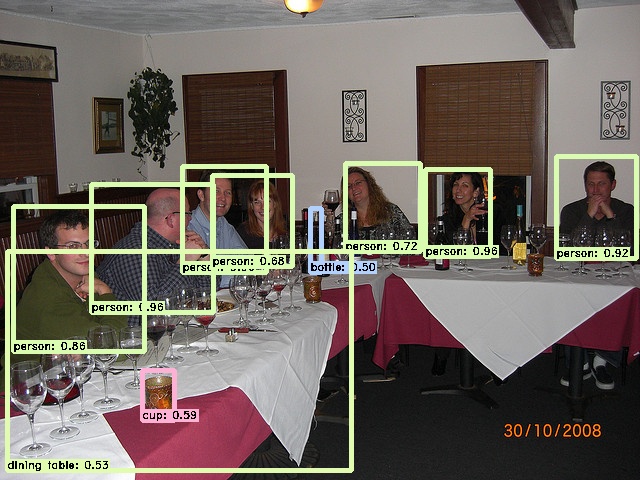} \\
        \includegraphics[height=0.255\linewidth]{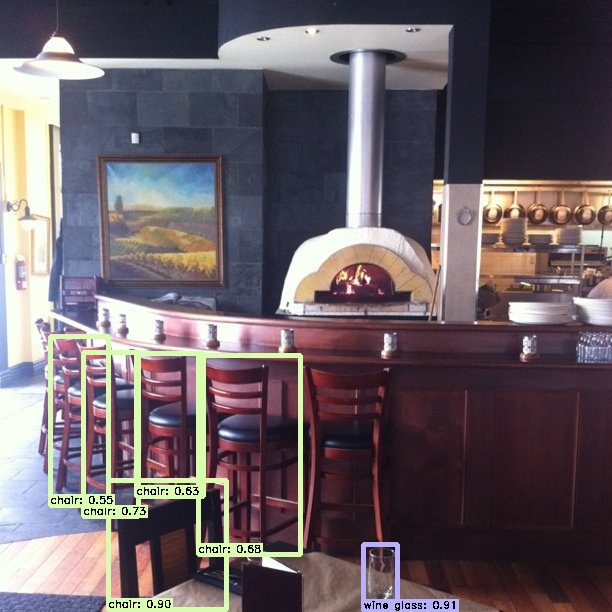}
        \hfill
        \includegraphics[height=0.255\linewidth]{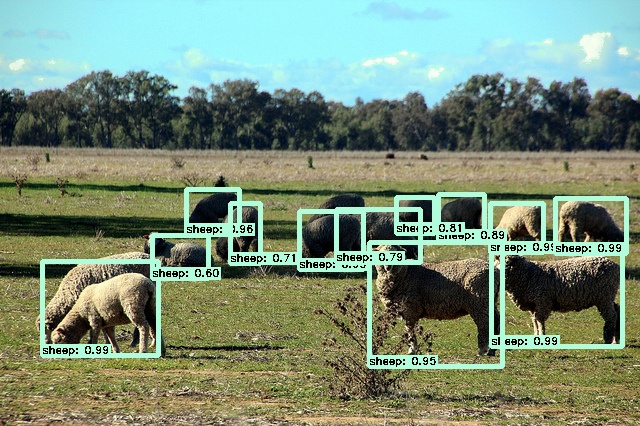}
        \hfill
        \includegraphics[height=0.255\linewidth]{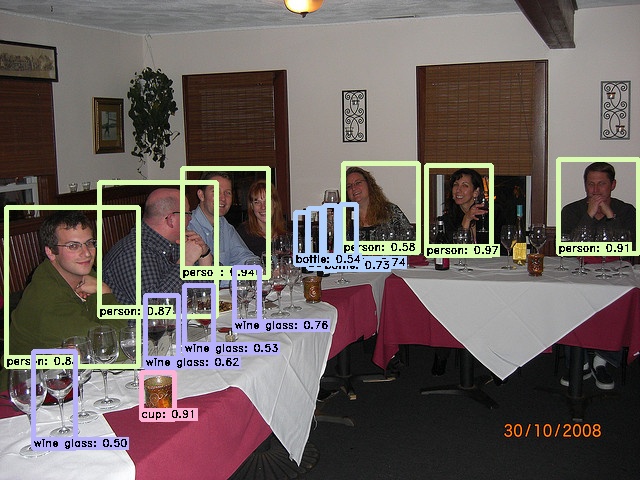}        \\
        \vspace{-0.5cm}
        \end{subfigure}
\end{center}
\caption{
\textbf{Detection examples on group of objects}. SSD does not detect the group of objects well, but our method works well.}
\label{fig:add_figures2}
\end{figure*}

\begin{figure*}[ht]
\begin{center}
	\begin{subfigure}{\textwidth}    	
	    \includegraphics[height=0.243\linewidth]{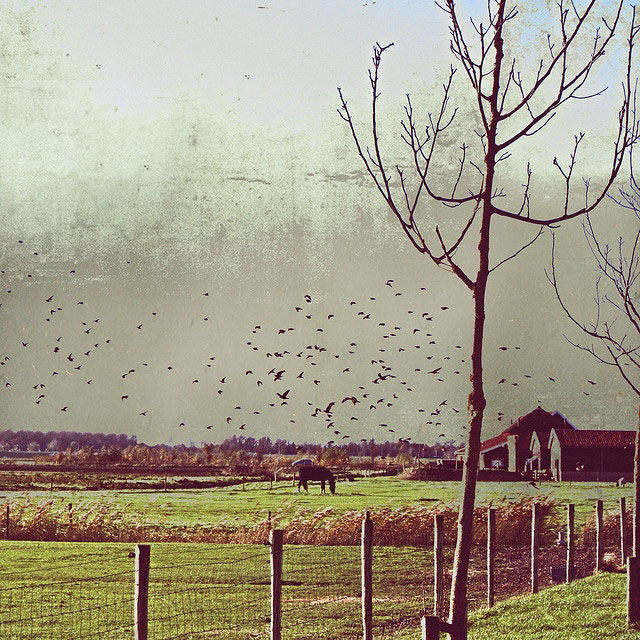}       		\hfill
        \includegraphics[height=0.243\linewidth]{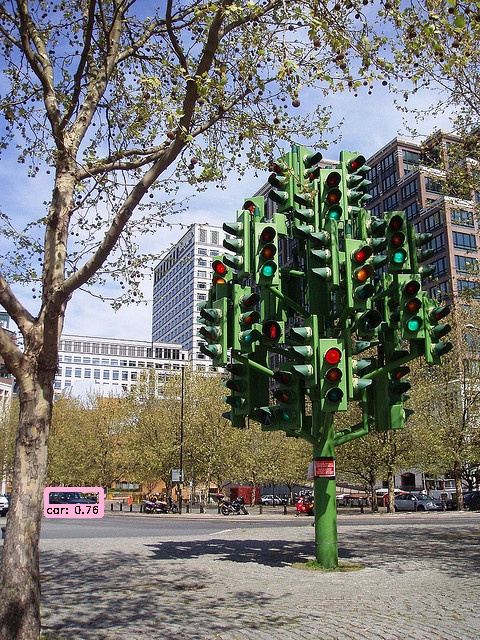}
\hfill
        \includegraphics[height=0.243\linewidth]{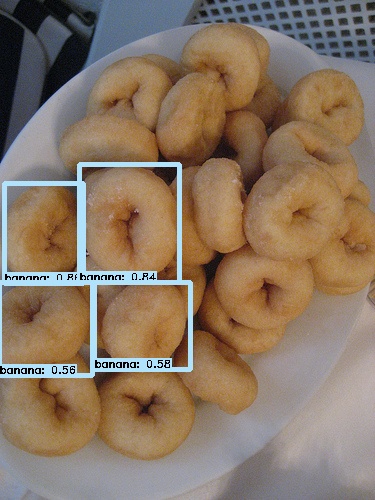}
        \hfill
        \includegraphics[height=0.243\linewidth]{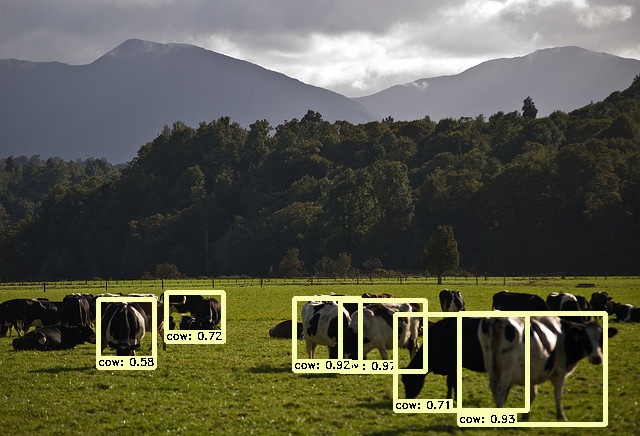} \\
        \includegraphics[height=0.243\linewidth]{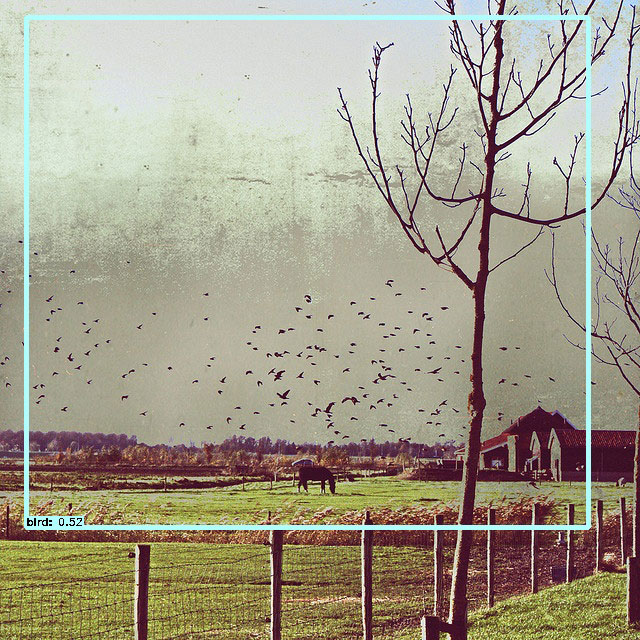}
        \hfill
        \includegraphics[height=0.243\linewidth]{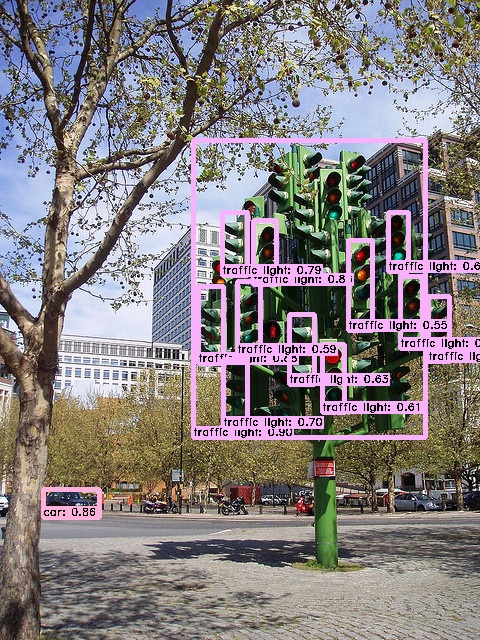}
        \hfill
        \includegraphics[height=0.243\linewidth]{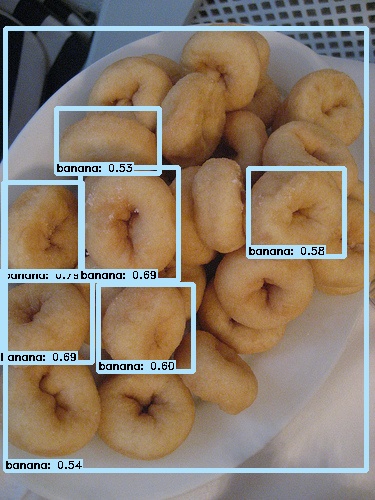}
        \hfill
        \includegraphics[height=0.243\linewidth]{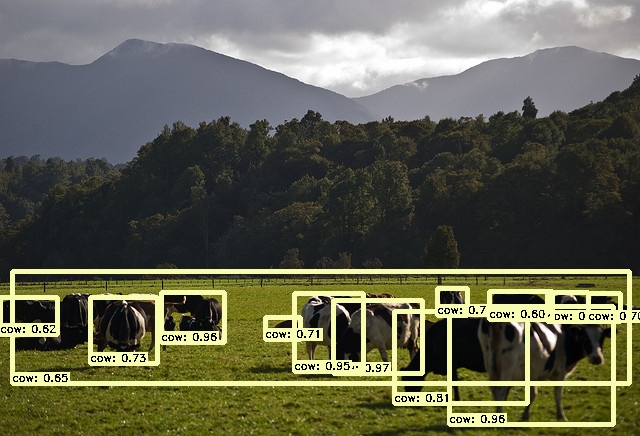} \\
        \vspace{-0.5cm}
        \end{subfigure}
\end{center}
\caption{
\textbf{Incorrect detection examples}. Our method often detects a large box which covers a group of objects.}
\label{fig:add_figures3}
\end{figure*}

\end{document}